\documentclass[conference]{IEEEtran}
\IEEEoverridecommandlockouts
\usepackage{cite}
\usepackage{amsmath,amssymb,amsfonts,multirow,}
\usepackage{graphicx}
\usepackage[table,xcdraw]{xcolor}
\usepackage{textcomp}
\usepackage{xcolor}
\usepackage{ctable}
\usepackage{subcaption}
\usepackage{soul}
\usepackage[ruled,vlined,linesnumbered]{algorithm2e}
\usepackage{algpseudocode}

\usepackage{authblk}

\def\BibTeX{{\rm B\kern-.05em{\sc i\kern-.025em b}\kern-.08em
 T\kern-.1667em\lower.7ex\hbox{E}\kern-.125emX}}
\begin{document}

\title{A Safe Deep Reinforcement Learning Approach for Energy Efficient Federated Learning\\ in Wireless Communication Networks}

\author{Nikolaos Koursioumpas*, Lina Magoula*, Nikolaos Petropouleas*,\\Alexandros-Ioannis Thanopoulos*, Theodora Panagea*, Nancy Alonistioti*\\ 
M. A. Gutierrez-Estevez**, Ramin Khalili**
\\
* \emph{Dept. of Informatics and Telecommunications, National and Kapodistrian University of Athens, Greece} 
\\
** \emph{Munich Research Center, Huawei Technologies Duesseldorf GmbH, Munich, Germany}
\\

\{nkoursioubas*, lina-magoula*, nipet*, gthanopoulos*, dpanagea*, nancy*\}@di.uoa.gr \\

\{m.gutierrez.estevez**, ramin.khalili**\}@huawei.com}

\maketitle

\begin{abstract}
Progressing towards a new era of Artificial Intelligence (AI) - enabled wireless networks, concerns regarding the environmental impact of AI have been raised both in industry and academia. Federated Learning (FL) has emerged as a key privacy preserving decentralized AI technique. Despite efforts currently being made in FL, its environmental impact is still an open problem. Targeting the minimization of the overall energy consumption of an FL process, we propose the orchestration of computational and communication resources of the involved devices to minimize the total energy required, while guaranteeing a certain performance of the model. To this end, we propose a Soft Actor Critic Deep Reinforcement Learning (DRL) solution, where a penalty function is introduced during training, penalizing the strategies that violate the constraints of the environment, and contributing towards a safe RL process. A device level synchronization method, along with a computationally cost effective FL environment are proposed, with the goal of further reducing the energy consumption and communication overhead. Evaluation results show the effectiveness and robustness of the proposed scheme compared to four state-of-the-art baseline solutions on different network environments and FL architectures, achieving a decrease of up to 94\% in the total energy consumption.
\end{abstract}


\section{Introduction}\label{Intro}
The Information and Communication Technology (ICT) industry represents an important energy consumer, utilizing 4\% of the world’s
electricity \cite{LANGE2020106760}. Progressing from 5G to Beyond 5G (B5G) and 6G networks, it becomes apparent that as the density of the network
infrastructure along with the number of connected devices exponentially increases, so does the energy demand. The ICT industry is expected to reach 10-20\% of the world’s
electricity by 2030 \cite{global5g},\cite{challe6010117}. Many ambitious climate change activities are already taking place by the European Commission across the ICT industry, in line with the GSMA’s commitment towards net-zero carbon emissions by 2050 \cite{GSMA_1, EU_Regulations}.  

In future wireless networks, the use of Artificial Intelligence (AI) as a key enabler has been recognized by both industry and academia \cite{kaloxylos_alexandros_2020_4299895}. Centralized AI approaches require a vast amount of data transfer to centrally located and energy hungry data centers, raising concerns with regard to both data privacy and energy consumption \cite{Savazzi9807354}. Alternative solutions such as Federated Learning (FL) have surfaced, where devices perform a model training in a collaborative manner, exploiting locally stored datasets and avoiding any raw data transmissions. FL could be effectively deployed to numerous use cases and applications, where indicative examples include smart cities, smart transportation and Industrial Internet of Things (IIoT) \cite{Shaheen11040670,Li_2020,ETSI-PDL}. Towards this direction, the 3rd Generation Partnership Project (3GPP) in Rel. 18, tries to adopt FL in Network Data Analytics Function (NWDAF) \cite{3GPP-FL}, with the introduction of the Model Training Logical Function (MTLF) in Rel. 17 that is responsible of training Machine Learning (ML) models and exposing new training services \cite{ETSI-MTLF}. 

Several critical challenges have been raised related to the application of FL to wireless networks, significantly affecting the overall energy consumption of the involved devices \cite{Niknam9141214}. Firstly, the model training process requires constant model update transmissions from the computing devices. Such model updates could consist of complex neural networks comprising millions or even billions of parameters. This implies a significant communication overhead, making the uplink transmissions particularly challenging. On the one hand, from the computing device perspective, transmitting a large volume of model updates periodically, imposes notable energy demands. On the other hand, the communication channel occupancy increases proportionally to the number of computing devices. The smooth operation of an FL process may be significantly impacted, particularly in the case of dynamic network environments with time-varying network conditions (e.g constantly changing channel conditions, fluctuations in traffic congestion). Secondly, ML model architectures require complex calculations, tightly related to the size of the ML model, resulting in even more challenges in terms of the energy aspect, since significant computational work will take place on device level during training. Another critical challenge of major importance includes computation and data heterogeneity across devices, in conjunction with strict ML model performance requirements. More specifically, critical services and applications often require stringent ML model performance \cite{5GCroCo_D2_1}, which is directly translated to higher training times and as a result to an increase in the overall energy consumption. Such deterioration could be accelerated in case of devices differing in both available resources and statistical data distribution. 

Taking into consideration the aforementioned challenges, it becomes apparent that new solutions are required to mitigate the environmental impact of FL in dynamic wireless communication networks, while guarantying a certain ML model performance. This could be achieved through the reduction of the energy consumption required to complete all the FL-related tasks. A number of state-of-the-art works focus on energy efficiency, by examining the problem in terms of resource orchestration, computation offloading or load balancing strategies (e.g. \cite{9545925Wang,8488502Cao,9145118Zeng}). However, only a limited number propose resource orchestration in run time (e.g. \cite{9569716Zhang,9384231Zhan}) and to the best of our knowledge only one work considers the complexity of an FL model architecture \cite{9475121Mo}. In addition, none of the existing works propose a safe learning approach that will enhance the decision making by selecting strategies that respect the constraints of the environment. In parallel, there are no solutions considering the potentially prohibited energy cost that is required, while exploring for optimal resource orchestration strategies towards energy efficient FL processes. 

The current work targets the minimization of the overall energy consumption of an FL process in a wireless communication network, by orchestrating the computational and communication resources of the involved devices, while guaranteeing a certain FL model performance target. The FL model transmissions are taking place in a synchronized manner towards eliminating the effect of straggling devices. This synchronization is realized at the device level, so as to avoid unnecessary FL model transmissions due to time overdue, and as a result, contribute not only to the reduction of the overall energy consumption but also to the overall communication channel occupancy. Finally, in dynamic wireless communication networks, the heterogeneity characterizing both the devices and the data varies over time, hence up-to-date decisions should be made in a run-time manner. 

In order to achieve a run-time decision making, we propose a Deep Reinforcement Learning (DRL) solution, utilizing the Soft Actor-Critic (SAC) algorithm. SAC is known to achieve efficient learning, stability and robustness \cite{haarnoja18b}. A SAC agent is trained under different conditions of the environment in an offline phase, in order to produce a robust model that is used in a run-time inference phase, orchestrating the computational and communication resources of the devices that are involved in an FL process. The key contributions can be summarized as follows:

\begin{itemize}

  \item We address the problem of FL energy efficiency in a wireless communication network, considering the model complexity, device and data heterogeneity, while guaranteeing a pre-selected FL performance target.
  \item We formulate a joint computation and transmission optimization problem aiming to minimize the total energy consumption of all devices involved in the FL process, while mitigating any wasted resource utilization.
  \item Considering dynamic environments, we propose a DRL solution utilizing the Soft Actor-Critic (SAC) algorithm to solve the optimization problem in a run-time manner. By introducing a penalty function to the training phase, we highly contribute towards a safe RL process \cite{2886795Garc}.
  \item We propose a device level synchronization in an attempt to eliminate unnecessary FL model transmissions, reducing energy consumption and communication overhead. 
  \item In order to avoid any excessive energy costs for the training phase of SAC, we propose a computationally cost effective FL environment, emulating a real FL process.
\end{itemize}

The rest of the paper is organized as follows. Section \ref{related_work} presents relevant state-of-the-art works. Section \ref{system_model} provides the system model. Section \ref{problem_formulation} provides the problem formulation. Section \ref{proposed_solution} describes the proposed solution that is evaluated in Section \ref{performance_evaluation} using the simulation setup of Section \ref{simulation_setup}. Finally, section \ref{conclusions} concludes the paper.

\vspace{-3pt}
\section{Related Work}\label{related_work}
\vspace{-3pt}
Over the recent years, energy efficiency in wireless communication networks has been the primary focus for a number of research studies, addressing the problem from different perspectives through the exploitation of resource orchestration, computation offloading, as well as load balancing strategies. 

A subset of research works is tackling the problem of energy efficiency, considering generic (non FL related) computation tasks \cite{8488502Cao,9760005Zhang,9500265Rezazadeh,9462445Zhou,8713902AlQerm}. In \cite{8488502Cao}, the authors consider two computation offloading schemes for Mobile Edge Computing (MEC) systems, improving the energy efficiency for latency-constrained computation by optimizing the available resources. Zhang \textit{et al.} in \cite{9760005Zhang} introduce an improved SAC algorithm towards a joint optimization of partial task offloading and resource allocation decisions. Another work \cite{9500265Rezazadeh} presents a knowledge plane-based MANO framework, using a twin-delayed double-Q SAC method towards energy consumption and Virtual Network Function instantiation cost minimization. In \cite{9462445Zhou}, the authors propose a Q-learning and Double Deep Q Networks (DDQN)-based methods to determine the joint policy of computation offloading and resource allocation in a dynamic MEC system. AlQerm \textit{et al.} in \cite{8713902AlQerm}, propose an online RL methodology to determine the most energy efficient traffic offloading strategy.  

There is a number of proposed solutions that consider an FL process as a use case, targeting to address energy-related challenges \cite{9545925Wang,9473734Ren,9475121Mo,9145118Zeng,Yang13037554,9569716Zhang,9384231Zhan,WANG2022199,9221089Nguyen}. The work in \cite{9545925Wang} presents a framework that targets to minimize the overall energy consumption of a Federated Edge Intelligence -supported IoT network, using the Alternate Convex Search algorithm. The authors in \cite{9473734Ren} propose a  joint resource allocation scheme for FL in IoT, aiming at the minimization of the system and learning costs by jointly optimizing bandwidth, computation frequency, transmission power allocation and sub-carrier assignment. Mo \textit{et al.} \cite{9475121Mo} focus on minimizing the total energy consumption of an FL system, by optimizing both communication and computational resources using techniques from convex optimization. Another work \cite{9145118Zeng}, proposes energy-efficient strategies for bandwidth allocation and scheduling, so as to reduce the energy consumption, while warranting learning performance in an FL framework. In \cite{Yang13037554} a bisection-based algorithm is proposed, whose goal is to minimize the total energy consumption of an FL system under a latency constraint. Zhang \textit{et al.} \cite{9569716Zhang}  present an energy-efficient FL framework for Digital Twin-enabled IIoT that exploits a DDQN, in order to jointly optimize training strategies and resource allocation, considering a dynamic environment. Another work \cite{9384231Zhan}, through the use of a Proximal Policy Optimization-based actor-critic method, targets the energy efficiency improvement of FL, by jointly minimizing the learning time and energy consumption. The authors in \cite{WANG2022199} exploit the merits of Multi-Agent Deep Deterministic Policy Gradient in resource allocation, addressing the challenges of FL in an Internet of Vehicles (IoV) scenario. Nguyen \textit{et al.} \cite{9221089Nguyen} propose a Deep Q Learning (DQL) algorithm concerning the resource allocation, in a mobility-aware FL network, which aims to maximize the number of successful transmissions, while minimizing the energy and channel costs. The work \cite{fliidopt} targets to optimize the performance of an FL model, using multi-agent RL, while considering energy-related requirements. The authors in \cite{fliotopt} propose an RL solution for device selection at each communication round, towards FL performance optimization.

It must be noted that only a limited amount of the state-of-the-art works \cite{9760005Zhang,9500265Rezazadeh,8713902AlQerm,9569716Zhang,9384231Zhan} propose solutions that can be deployed in a run-time manner, taking into account the dynamicity of wireless network environments. Also, only the authors in \cite{9475121Mo} take into account the FL model's complexity in conjunction with its size that could significantly affect the total energy consumption of the system. Moreover, there are no related works incorporating the wasted amount of consumed energy for any unfinished FL task (e.g. due to time limit) as part of the solution. Also, none of the existing RL works  consider the energy cost that is required, while exploring for energy efficient strategies. To the best of our knowledge, our paper is the first work that addresses the minimization of the overall FL related energy consumption and communication channel occupancy in a run-time and safe \cite{2886795Garc} manner, tailored to any time-varying wireless network environment, while taking into account the FL model's performance and complexity. Finally, this work towards addressing the energy cost of the offline RL phase, proposes a computationally cost effective FL environment that emulates a real FL process. Table \ref{tab:related_work} provides a brief summary of the related work.
\begin{table}[ht]
  \centering
  \begin{tabular}{|p{0.1\textwidth}|p{0.3308\textwidth}|} \hline 
       \textbf{Reference}& \textbf{Main Contribution}\\ \hline 
       \cite{8488502Cao,9760005Zhang,9500265Rezazadeh,9462445Zhou,8713902AlQerm} & Energy efficiency in generic computations (non-FL)\\ \hline 
       \cite{9545925Wang,9473734Ren,9475121Mo,9145118Zeng,Yang13037554,9384231Zhan} & Non-RL solutions for energy efficiency in FL\\ \hline 
       \cite{9569716Zhang, 9221089Nguyen, fliidopt} & RL solutions for energy efficiency in FL\\ \hline 
\cite{WANG2022199, fliotopt} & Non-energy aware RL solutions for FL optimizations\\\hline
  \end{tabular}
  \caption{Summary of Related work}
  \label{tab:related_work}
  \vspace{-3pt}
\end{table}

\section{System Model} \label{system_model}
\vspace{-3pt}
This section provides the system model of an AI-enabled wireless communication network using FL. The set of workers, denoted by $\mathcal{K}$, is orchestrated by a coordinator node. All workers, during an FL process, contribute to the training of a Neural Network (NN).  

An FL process is comprised of a number of rounds. For the rest of the paper, each round is named as \textit{global} iteration and it is denoted by $n \in \mathbb{N}$. Let $\bold{w}_n \in \mathbb{R}$ be a vector containing the model parameters at \textit{global} iteration $n$. In each iteration $n$, the coordinator node distributes the $\textbf{w}_n$ of a global NN model of size $m \in \mathbb{R}_+$, in bits, to the workers that are involved in the FL process. Let $\alpha \in \mathbb{R}_+$ denote the complexity associated with the global model and measured in Floating Point Operations (FLOPs). Each worker $k \in \mathcal{K}$, after successfully receiving the vector with the global parameters $\textbf{w}_n$, performs a local training using its locally stored dataset $\mathcal{D}_{k,n}$ that includes also the set of groundtruth data $\mathcal{L}_{k,n}$ ($\mathcal{L}_{k,n} \subset \mathcal{D}_{k,n}$). We denote the variance of the dataset as $v_{k,n}$, expressing the degree of spread in the groundtruth data, given by Eq. (\ref{variance_dataset}). 
\begin{equation} \label{variance_dataset}
  v_{k,n} = \mathrm{Tr}{\left(\frac{\sum_{i=1}^{s_{k,n}}({\mathcal{L}}_{k,n,i} - \overline{{\mathcal{L}}}_{k,n})^\mathrm{T}({\mathcal{L}_{k,n,i}} - \overline{\mathcal{L}}_{k,n})}{s_{k,n}-1}\right)},
\end{equation}
where $\mathcal{L}_{k,n,i}$ and $\overline{{\mathcal{L}}}_{k,n}$ denote the $i^{th}$ row of $\mathcal{L}_{k,n}$ and the mean vector of the columns of the groundtruth data, respectively. As $s_{k,n} \in \mathbb{N}$ we denote the total number of samples of $\mathcal{D}_{k,n}$. The $\mathrm{Tr}$ denotes the trace of a matrix, which is equal to the sum of its diagonal elements.

\textbf{Device Computational Capability}: Let $f_{k,n} \in \mathbb{R}_+$ denote the available computational capacity (i.e. Central Processing Unit (CPU) speed) of worker $k$ to execute a local training at the $n^{th}$ \textit{global} iteration\footnote{It is assumed that the coordinator node has enough computational capacity to execute any computational task}. Based on the available computational capacity $f_{k,n}$, the worker $k$ can complete a certain number of FLOPs per cycle, denoted by $c_{k,n} \in \mathbb{R}_+$. The CPU of each worker $k$ has also an effective switched capacitance, denoted by $\varsigma_k \in \mathbb{R}_+$, which depends on the hardware architecture. The local training of each worker is considered complete when a pre-selected performance target $\eta \in [0,1]$ is reached. The pre-selected performance target $\eta$ represents a threshold that is imposed to the local training of all workers and could refer to error metrics (e.g. mean squared error (MSE), cross entropy loss), accuracy, etc. Let $I_{k,n} \in \mathbb{N}$ be the total number of \textit{local} iterations (local training rounds) required in order for the locally trained NN model of worker $k$ to reach the pre-selected performance target $\eta$ at \textit{global} iteration $n$. By $\tau_{k,n} \in \mathbb{R}_+$ we denote the time required by worker $k$ to complete a local training process.

\textbf{Device Communication Capability: } After $I_{k,n}$ \textit{local} iterations, each worker $k$ produces a locally trained NN model with parameters $\textbf{w}_{k,n}$ that needs to be transmitted to the coordinator. The communication channel between worker $k$ and the coordinator at \textit{global} iteration $n$ is modeled as a flat-fading channel with Gaussian noise power density $N_0 \in \mathbb{R}$ and channel gain $g_{k,n} \in \mathbb{R}$, where the fading is assumed constant during the transmission of the model. Also, let $b_{k,n} \in \mathbb{R}_+$ and $p_{k,n} \in \mathbb{R}_+$ be the assigned bandwidth and transmission power to worker $k$ at \textit{global} iteration $n$, respectively\footnote{We assume that enough bandwidth has been assigned to $k$, in order to transmit its model updates.}. By $r_{k,n} \in \mathbb{R}_+$ and $t_{k,n} \in \mathbb{R}_+$ we denote the achievable data rate and the required time to upload $\textbf{w}_{k,n}$ to the coordinator, respectively.

\textbf{FL Rounds: } The FL process is realized in a synchronized manner. A \textit{global} iteration is finished when the coordinator receives updates from all workers or when a pre-selected time threshold $\mathsf{H} \in \mathbb{R}_+$ is reached. The selection of $\mathsf{H}$ is tailored to the use case and is imposed by the coordinator. Workers should complete their computational and communication tasks within this time threshold. Updates from those workers who have not met the time threshold are considered invalid and are not used to update the global model, and thus their used energy is deemed as wasted. Workers with zero assigned computational and/or communication resources (i.e. workers that are not selected for this round of training -as it will be shown later on Sec. \ref{proposed_solution}) are excluded from this time threshold related condition. The coordinator node produces an updated global model $\textbf{w}_{n+1}$ using all the received model updates following the Federated Averaging (FedAvg) technique \cite{Li_2020} as provided by Eq. (\ref{global_ML_parameters}), and broadcasts it to the workers before they start the next \textit{global} iteration ($n+1$). This procedure is repeated until the global NN model reaches a pre-selected performance target $\epsilon_0 \in [0,1]$. Once again, this performance target could refer to error metrics, accuracy etc. 
\vspace{-7pt}
\begin{equation} \label{global_ML_parameters}
  \textbf{w}_{n+1} = \frac{1}{v}\sum_{k=1}^{\mathcal{K}}v_{k,n}\cdot \textbf{w}_{k,n},
  \vspace{-6pt}
\end{equation}

where
\vspace{-4pt}
\begin{equation} \label{variance_weights}
  v = \sum_{k=1}^{\mathcal{K}}v_{k,n}.
  \vspace{-4pt}
\end{equation}

Table \ref{table_notations} summarizes the notations. It must be noted that the above-described process is applicable to a wider part of the wireless communication network, where multiple coordinator nodes  span across different layers, following an hierarchical structure. As a result, each coordinator node, apart from orchestrating the associated workers, can cooperate with other respective coordinator nodes towards contributing to the NN model parameters' aggregation processes of higher layers in the hierarchy. 

\vspace{-5pt}
\begin{table}[ht]
 \renewcommand{\arraystretch}{1.05}
 \begin{tabular}{|p{0.048\textwidth}|p{0.39\textwidth}|}
\hline
\textbf{Params} & \textbf{Description} \\
\hline
$\mathcal{K}$ & Set of workers \\ \hline
$n$ & Index of the \textit{Global} iteration (FL round) \\ \hline
$N_0$ & White Gaussian noise power spectral density\\ \hline
$g_{k,n}$ & Gain of the wireless channel the worker $k$ has access to at the $n^{th}$ \textit{global} iteration\\ \hline
$f_{k,n}$ & Available computational capacity of worker $k$ at the $n^{th}$ \textit{global} iteration\\ \hline
$b_{k,n}$ & Bandwidth assigned to worker $k$ at the $n^{th}$ \textit{global} iteration\\ \hline
$p_{k,n}$ & Transmission power of worker $k$ at the $n^{th}$ \textit{global} iteration \\ \hline
$r_{k,n}$ & Achievable transmission data rate of worker $k$ at the $n^{th}$ \textit{global} iteration\\ \hline
$\mathcal{D}_{k,n}$ & Local dataset of worker $k$ at the $n^{th}$ \textit{global} iteration\\ \hline
$\mathcal{L}_{k,n}$ & The set of groundtruth data of worker $k$ at the $n^{th}$ \textit{global} iteration\\ 
\hline
$s_{k,n}$ & Total number of data samples of worker $k$ at the $n^{th}$ \textit{global} iteration\\ \hline
$v_{k,n}$ & Dataset variance of worker $k$ at the $n^{th}$ \textit{global} iteration\\ \hline
$c_{k,n}$ & Total number of Floating Point Operations (FLOPs) per cycle that the worker $k$ can complete at the $n^{th}$ \textit{global} iteration\\ \hline
$\varsigma_k$ & Effective switched capacitance of worker $k$ \\ \hline
$\textbf{w}_n$ & Global FL model produced at the $n^{th}$ \textit{global} iteration \\ \hline
$\alpha$ & Complexity of the global FL model in terms of total number of Floating Point Operations (FLOPs) \\ \hline
$m$ & Size of global FL model in bits \\ \hline
$I_{k,n}$ & Number of \textit{local} iterations required to reach $\eta$ at the worker $k$ at the $n^{th}$ \textit{global} iteration\\ \hline
$\tau_{k,n}$ & The time required by a worker $k$ to complete a local training process at the $n^{th}$ \textit{global} iteration\\ \hline 
$t_{k,n}$ & The time required by worker $k$ to transmit its model updates at the $n^{th}$ \textit{global} iteration\\ \hline
$\textbf{w}_{k,n}$ & Model parameters of worker $k$ at the $n^{th}$ \textit{global} iteration \\ \hline
$\eta$ & Training performance target of all workers\\ \hline
$\epsilon_0$ & Training performance target of the global FL model\\ \hline
\end{tabular}
\caption{Notation Table} 
\label{table_notations} 
\vspace{-15pt}
\end{table}

\section{Problem Formulation} \label{problem_formulation}
 The objective of our problem formulation  is to achieve energy efficiency in the system, i.e the minimization of the workers' overall energy consumption, while guarantying a certain global FL model performance target. All workers consume a specific amount of energy from the power grid, denoted by $E_{gr}$, in order to complete an FL process. Two types of tasks are considered in an FL process, the computation tasks, referring to the local training processes, and the communication ones, related to the transmission of the model parameters. Our optimization problem is  modeled  as  a  Markov Decision Process (MDP) \cite{Wiley1994},  with  the  aim of maximizing the cumulative discounted rewards in the long term. The definitions of the sets of states $\mathcal{S}$, actions $\mathcal{A}$ and rewards $\mathcal{R}$ are provided in Section \ref{proposed_solution}.

 In order to avoid any unnecessary transmissions due to time overdue, the synchronization is performed at the worker side. If a worker does not manage to meet the time threshold, it discards its model updates, performing no transmission. This could happen if the worker requires higher number of local iterations than initially anticipated to reach the local performance target $\eta$. The elimination of any unnecessary transmissions contributes to the reduction of the overall energy consumption and communication channel occupancy, at the expense of wasting the energy used for computation of those workers who do not meet the time threshold. The computation energy considered as wasted is denoted by $E^W_{k,n}$.

The amount of computation energy required by a worker $k$ to perform a computation task at the $n^{th}$ \textit{global} iteration is denoted by $E_{k,n}^C$ and computed as follows: 
\vspace{-4pt}
\begin{equation}  \label{computation_energy}
  E_{k,n}^C = \frac{\varsigma_{k} \cdot I_{k,n} \cdot \alpha \cdot s_{k,n} \cdot f_{k,n}^2} {c_{k,n}}.
  \vspace{-2pt}
\end{equation}
The amount of transmission energy required by a worker $k$ to perform a communication task at the $n^{th}$ \textit{global} iteration is denoted by $E_{k,n}^T$ and computed as: 
\vspace{-4pt}
\begin{equation} \label{transmission_energy}
  E_{k,n}^T = \frac{m \cdot p_{k,n}}{r_{k,n}},
  \vspace{-4pt}
\end{equation}
where
\vspace{-4pt}
\begin{equation}
  r_{k,n} = {b_{k,n} \cdot log_2\left(1 + \frac{g_{k,n} \cdot p_{k,n}}{b_{k,n} \cdot N_0}\right)}.
  \vspace{-2pt}
\end{equation}
 The two vectors to be optimized are the $\textbf{f}_n =[f_{1,n},...,f_{\mathcal{K},n}]^T$and $\textbf{p}_n = [p_{1,n},...,p_{\mathcal{K},n}]^T$, representing the computational capacity and the transmission power of all workers at each \textit{global} iteration $n$ of the FL process. 

The objective of our problem formulation is defined as:
\vspace{-5pt}
\begin{equation} \label{objective_function}
  \min_{ \textbf{f}_n, \textbf{p}_n} E_{gr} =  \sum_{n=1}^{I_0} \gamma^{n-1}\sum_{k=1}^{\mathcal{K}} (E_{k,n}^C\cdot \Omega_{k,n} + E_{k,n}^T)
\end{equation}
\vspace{-17pt}
\begin{eqnarray} \label{complete_version_constraint_total_latency}
    \textrm{s.t.}&\tau_{k,n} + t_{k,n} < \mathsf{H}, \forall k \in \mathcal{K} 
    \vspace{-10pt}
\end{eqnarray}
\vspace{-18pt}
\begin{eqnarray} && \label{complete_version_constraint_per_device_capacity}
  0\leq f_{k,n} \leq f_{k,n}^{max}, \forall k \in \mathcal{K} 
  \vspace{-4pt}
\end{eqnarray}
\vspace{-20pt}
\begin{eqnarray} &&
\label{complete_version_constraint_per_device_power}
  0 \leq p_{k,n} \leq p_{k,n}^{max} ,  \forall k \in \mathcal{K}
\end{eqnarray}
\vspace{-20pt}
\begin{eqnarray} &&
\label{complete_version_constraint_idle}
 \sum_{k=1}^{\mathcal{K}} f_{k,n} > 0,
\end{eqnarray}
\vspace{-8pt}
where:
\begin{equation} \label{computation_time}
  \tau_{k,n}= \frac{I_{k,n} \cdot \alpha \cdot s_{k,n}}{c_{k,n} \cdot f_{k,n}}, 
  \vspace{-5pt}
\end{equation}

\begin{equation} \label{transmission_time}
  t_{k,n}= \frac{m}{r_{k,n}}.
  \vspace{-5pt}
\end{equation}

More precisely, $I_0$ denotes the terminal state, i.e. the \textit{global} iteration in which the FL model reaches the pre-selected performance target $\epsilon_0$, $\gamma \in [0,1]$ is the discount rate to account for the relative importance of the energy consumption of future \textit{global} iterations, and
 $\Omega_{k,n}$, given in Eq. (\ref{indicator_function_omega}), is an indicator function ensuring that there will be no local training at  worker $k$ at the iteration $n$, in case of zero allocated transmission power.
 \vspace{-7pt}
\begin{equation} \label{indicator_function_omega}
  \Omega_{k,n} = 
   \left\{
  \begin{array}{ll}
  1, & p_{k,n} > 0\\
  0, & otherwise\\
  \end{array}
  \right.
  \vspace{-4pt}
\end{equation}
 Based on Eq. (\ref{transmission_energy}), in case of zero allocated computational capacity, there will be no model parameters for transmission, and as such the transmission energy will be also equal to zero. 
 
Constraint (\ref{complete_version_constraint_total_latency}) ensures the synchronization of the FL process at the worker side, by upper bounding the total time required by each worker to complete a computation and transmission task at the pre-selected time threshold $\mathsf{H}$. Constraints (\ref{complete_version_constraint_per_device_capacity}) and (\ref{complete_version_constraint_per_device_power}) ensure that the computational capacity along with the transmission power of each worker $k$ at the $n^{th}$ \textit{global} iteration are within the maximum available computation and communication capabilities of the worker, denoted by $f_{k,n}^{max}$ and $p_{k,n}^{max}$, respectively. Finally, constraint (\ref{complete_version_constraint_idle}) ensures that at least one worker should be involved in the FL process, contributing to the global model update.  

In order for the locally trained model to achieve a certain performance, the following condition is imposed to each worker $k$ at each \textit{global} iteration $n$ \cite{Yang13037554}:
 \vspace{-4pt}
\begin{equation} \label{local_accuracy_inequality}
  \frac{F_k(\textbf{w}_{k,n}^{(i)},\mathcal{D}_{k,n}) - F^*}{F_k(\textbf{w}_{k,n}^{(0)},\mathcal{D}_{k,n}) -  F^*} \leq \eta,
  \vspace{-4pt}
\end{equation}
where $F_k$ denotes the local objective function with regard to FL model performance of each worker $k$, $\textbf{w}_{k,n}^{(i)}$ denotes the local model parameters at the $n^{th}$ \textit{global} iteration at the $i^{th}$ \textit{local} iteration, and $F^*$ is the optimal value (minimum or maximum) of the global objective function $F$, tailored to the selected FL task. $F$ is resulted from the weighted average of the converged local objective functions of all $\mathcal{K}$ workers at the $n^{th}$ \textit{global} iteration, as follows:
\vspace{-5pt}
\begin{equation} \label{global_loss}
  F(\textbf{w}_{1,n},...,\textbf{w}_{\mathcal{K},n}) = \frac{1}{v}\sum_{k=1}^{\mathcal{K}}v_{k,n}\cdot F_k(\textbf{w}_{k,n},\mathcal{D}_{k,n}).
\vspace{-5pt}
\end{equation}
The left part of the condition (\ref{local_accuracy_inequality}) is referred as local performance rate and is denoted by ${e}_{k,n}^{(i)}$. In the same notion, the following performance condition is imposed to the global FL model at each \textit{global} iteration $n$, in order to converge to a pre-selected performance target $\epsilon_0$:
\vspace{-5pt}
\begin{equation} \label{global_accuracy_inequality}
  \frac{F(\textbf{w}_{1,n},...,\textbf{w}_{\mathcal{K},n}) - F^*}{F(\textbf{w}_{1,0},...,\textbf{w}_{\mathcal{K},0}) - F^*} \leq \epsilon_0.
\end{equation}
The left part of the condition (\ref{global_accuracy_inequality}) is referred as global performance rate and is denoted by ${e}_{n}$. The FL process stops, when the condition in (\ref{global_accuracy_inequality}) is satisfied. 
Given the fact that the computational capacity and transmission power of each worker must be selected at each \textit{global} iteration, it is intuitive that the problem should be solved in a run-time manner.
\vspace{-2pt}
\section{Proposed Deep Reinforcement Learning Solution} \label{proposed_solution}
RL is an exploration process aiming at maximizing a long-term reward through a sequence of interactions with an environment. In an RL process, there is an agent that monitors three sets of elements known as states $\mathcal{S}$, actions $\mathcal{A}$ and rewards $\mathcal{R}$. On each step of interaction $n$ the agent observes a state $S_{n} \in \mathcal{S}$ that is the current representation of the environment, and selects an action $A_{n} \in \mathcal{A}$. At the next step $n+1$, as a consequence of the selected action, the agent receives a reward $rr_{n+1} \in \mathcal{R}$ (i.e. a numerical feedback) and transitions to the next state of the environment $S_{n+1} \in \mathcal{S}$ \cite{10.5555/3312046}. We propose a SAC DRL solution to solve the optimization problem introduced in Section \ref{problem_formulation}. SAC is applicable to complex optimization problems with continuous solution space (continuous RL actions), as we consider in this paper. Contrary to other well-known RL algorithms for continuous action space \cite{Dutta24508}, SAC targets not only to maximize the long-term reward but also explicitly encourages the exploration of different interactions with the environment and avoids repeatedly selecting a particular action. Besides, SAC is easily tunable (less hyper-parameters) and promises faster convergence \cite{haarnoja18b}. 

 Our problem is considered as episodic, where each episode ends in a special state, named as terminal state and is determined by the FL model's convergence. The main definitions of the proposed SAC solution, are provided below:

\textbf{Environment: } The environment, that the SAC agent constantly monitors, is comprised of the coordinator node and the associated workers that are involved in an FL process. 

\textbf{RL step: } An RL step represents a \textit{global} iteration of an FL process. In each RL step the SAC agent monitors and interacts with the environment by assigning resources to the workers involved in the FL process.

\textbf{RL episode: } An RL episode represents a complete FL process. More specifically, an RL episode consists of a number of RL steps, equal to the total number of \textit{global} iterations of an FL process. Each RL episode terminates when the FL model reaches the pre-selected performance target $\epsilon_0$.

\textbf{RL State: } The RL state represents the information of the environment that the SAC agent monitors at each RL step. The RL state $S_{n}$ at step $n$ comprises the following information: the number of \textit{local} iterations at each worker ($I_{k,n-1}$), the wasted energy at each worker ($E^W_{k,n-1}$), the global performance rate ($e_{n-1}$), the maximum available computation and communication capabilities of each worker ($f_{k,n-1}^{max}$, $p_{k,n-1}^{max}$), the allocated bandwidth per worker ($b_{k,n-1}$), and the total number of data samples of each worker's local dataset ($s_{k,n-1}$) at RL step ${n-1}$. As a result, the state of the environment at the $n^{th}$ RL step is defined as:
\begin{equation*}
  S_{n} = \{\textbf{I}_{n-1}, \textbf{E}^W_{n-1}, e_{n-1}, \textbf{f}_{n-1}^{max}, \textbf{p}_{n-1}^{max}, 
  \textbf{b}_{n-1}, \textbf{D}_{n-1}\} ,
  \vspace{-5pt}
\end{equation*}
where: \\ $\textbf{I}_{n-1} = [I_{1,n-1},...,I_{\mathcal{K},n-1}]^T$, $\textbf{E}^W_{n-1} =
[E^W_{1,n-1},...,E^W_{\mathcal{K},n-1}]^T$, $\textbf{f}_{n-1}^{max} = [f_{1,{n-1}}^{max},...,f_{\mathcal{K},{n-1}}^{max}]^T$, $\textbf{p}_{n-1}^{max} = [p_{1,{n-1}}^{max},...,p_{\mathcal{K},{n-1}}^{max}]^T$,\\ $\textbf{b}_{n-1} = [b_{1,{n-1}},...,b_{\mathcal{K},{n-1}}]^T$, $\textbf{D}_{n-1} = [s_{1,{n-1}},...,s_{\mathcal{K},{n-1}}]^T$.

\textbf{Action Space: } The action space $A_{n}$, is comprised of all computational and communication resources assigned to all workers ($\textbf{f}_n$, $\textbf{p}_n$) by the agent at RL step $n$, i.e. 
\vspace{-4pt}
\begin{equation*}
  A_{n} = \{\textbf{f}_n, \textbf{p}_n\}. 
\vspace{-4pt}
\end{equation*}
Based on constraints (\ref{complete_version_constraint_per_device_capacity}) and (\ref{complete_version_constraint_per_device_power}), the continuous action space is bounded. The lower bounds (zero resource assignments) are specifically selected to enable a worker selection policy, potentially excluding workers with high energy consumption but limited contribution to the performance of the FL process.

\textbf{Reward Function: } The reward function of the SAC agent is formulated based on the objective function (\ref{objective_function}), in conjunction with constraints (\ref{complete_version_constraint_total_latency}) and (\ref{complete_version_constraint_idle}) (constraints (\ref{complete_version_constraint_per_device_capacity}) and (\ref{complete_version_constraint_per_device_power}) are satisfied by the bounded action space). The discount rate $\gamma$ in our case is set to $1$, since the optimization problem is episodic \cite{10.5555/3312046} and we want to assign the same importance to all retrieved future rewards. Specifically, the reward that the SAC agent receives at the $n^{th}$ \textit{global} iteration is defined as:
\vspace{-5pt}
\begin{equation}\label{sac_reward}
rr_{n} = - [ \sum_{k=1}^{\mathcal{K}} (E_{k,n}^C\cdot \Omega_{k,n} + E_{k,n}^T)+ x_{n}].
\vspace{-4pt}
\end{equation}

The first part of Eq. (\ref{sac_reward}) is the total energy consumption, i.e. computation and communication, of all workers that are involved in the FL process in the step $n$. The second part $x_{n}$ is a penalty term defined to contribute towards a safe RL process \cite{2886795Garc}. The SAC agent is penalized for the actions that resulted in wasted energy and violation of constraints (\ref{complete_version_constraint_total_latency}) and (\ref{complete_version_constraint_idle}), respectively. We define the penalty $x_{n}$ as follows:
\vspace{-5pt}
\begin{equation} \label{penalty}
  x_{n} = \sum_{k=1}^{\mathcal{K}}{(E_{k,n}^W + \mu_1 \cdot P^{(1)}_{k,n})} + \mu_2 \cdot P^{(2)}_{n},
\vspace{-5pt}
\end{equation}

where $E_{k,n}^W$ is the amount of wasted computation energy, as defined in Eq. (\ref{computation_energy}), in the case that the worker $k$ did not meet the time limit at the $n^{th}$ RL step. Let $\mu_1$ and $\mu_2$ denote the hyper-parameters weighing the impact of each constraint violation in the reward, and $P^{(1)}_{k,n}$ and $P^{(2)}_{n}$ are two indicator functions, related to constraints (\ref{complete_version_constraint_total_latency}) and (\ref{complete_version_constraint_idle}):
\vspace{-3pt}
\begin{equation} \label{device_constraint_violation_indicator_1}
  P_{k,n}^{(1)} = 
  \left\{
  \begin{array}{ll}
  0, & \tau_{k,n}+t_{k,n}-\mathsf{H}<0 \\
  1, & otherwise\\
  \end{array},
  \right.
\end{equation}
and 
\vspace{-5pt}
\begin{equation} \label{device_constraint_violation_indicator_2}
  P_{n}^{(2)} = 
  \left\{
  \begin{array}{ll}
  0, & \sum_{k=1}^{\mathcal{K}} f_{k,n} > 0\\
  1, & otherwise\\
  \end{array}.
  \right.
\end{equation}
\vspace{-10pt}

Taking into account Eq. (\ref{sac_reward}) and (\ref{penalty}), we can rewrite the reward as:
\vspace{-9pt}
\begin{equation}
rr_{n} = - [\sum_{k=1}^{\mathcal{K}}(E^C_{k,n}\cdot \Omega_{k,n} + E^T_{k,n} + E_{k,n}^W + \mu_1 \cdot P^{(1)}_{k,n}) + \mu_2 \cdot P^{(2)}_{n}].
\end{equation}

The SAC agent targets to maximize the reward, which results on minimizing the total energy consumption with respect to the constraints of the environment. 

\textbf{Offline Training and Online Inference: }The proposed solution consists of two main phases, the offline training phase of the SAC agent and the online inference phase. 
During the offline training phase (Algorithm \ref{offline_stage}), in each RL episode, the SAC agent monitors the environment, where an FL process is taking place. In each RL step, the SAC agent selects the appropriate actions $A_{n}$ (line \ref{enf_action}) that will be enforced to the workers in order to allow them to initialize local training processes (line \ref{init_local_train}). These processes complete, when the local performance target is reached (line \ref{local_errr_rate}). After receiving all the on-time transmissions from the respective workers (line \ref{tran_updates}), the coordinator proceeds with the model aggregation (line \ref{aggregate_updates}) and constructs the next state of the environment $S_{n+1}$ (line \ref{contruct_state}). For any constraint violations, the coordinator node adds the penalty $x_{n}$ to the reward $rr_{n}$ (line \ref{check_1}). The SAC agent stores the current state of the environment $S_{n}$, the calculated reward $rr_{n}$, the selected actions $A_{n}$, as well as the next state of the environment $S_{n+1}$ for training (line \ref{log_data}). This process is repeated until the global performance target is reached (line \ref{target_perf}). The final SAC model is exploited in the online inference phase (line \ref{final_model}), towards run-time decision making, since it does not require additional learning and as such logging of information \cite{Prudencio01387, Haarnoja05905}.

Despite the fast convergence of the SAC agent, the training procedure may require up to millions of episodes until convergence in dynamic environments. As it becomes apparent, the training process of the SAC agent, incorporating periodical executions of FL processes, is of high complexity and time consuming, but most importantly, contrary to the objective of the current work; it requires a significant amount of energy consumption. In order to overcome this issue and as it will be described in Section \ref{simulation_setup}, a significantly simpler FL environment was designed for the SAC training phase, emulating a real FL process for a given performance target. This is accomplished through the estimation of the necessary information that is related to the energy consumption of the system, i.e the number of \textit{local} iterations and the global performance rate achieved at each \textit{global} iteration. Through this simulated environment, the SAC agent can be efficiently trained, for a high number of episodes with different configurations, towards providing near optimal solutions. Finally, in the online inference phase and in order to evaluate the effectiveness and the robustness of the SAC agent, the trained SAC model is tested using real FL processes.
\begin{algorithm}
\small
\caption{Offline Training Phase SAC Agent}
\label{offline_stage}
  Initialize Network Environment\\
  Initialize and Configure the RL environment \\
  \ForEach{RL episode} {
    \ForEach{RL step $n$}{ 
    Select and enforce action $A_{n}$\\ \label{enf_action}
    \ForEach {$k \in \mathcal{K}$}{
      Initialize a local training process\\ \label{init_local_train}
      \ForEach {local iteration $i$}{
        Update $\textbf{w}^{(i)}_{k,n}$ \\
        \lIf{${e}_{k,n}^{(i)} \leq \eta$}{ \label{local_errr_rate}
        break
        }
      }
      \If{constraint (\ref{complete_version_constraint_total_latency}) is satisfied}{
        \textbf{Send}: \{$\textbf{w}_{k,n}$, $I_{k,n}$, $E^W_{k,n}$, $f^{max}_{k,n}$, $p^{max}_{k,n}$, $b_{k,n}$, $s_{k,n}$\} \label{tran_updates}
      }
    }
    Produce $\textbf{w}_n$ and calculate $e_n$ \\ \label{aggregate_updates}
    \textbf{Construct}: $S_{n+1} = \{\textbf{I}_n, \textbf{E}^W_n, e_n,\textbf{f}_{n}^{max}, \textbf{p}_{n}^{max}, \textbf{b}_{n}, \textbf{D}_{n}\}$ \\ \label{contruct_state}
    \ForEach {$k \in \mathcal{K}$}{ \label{violations_1}
      \lIf{violations}{ \label{check_1}
        update $x_{n}$  and apply to $rr_{n}$ \label{check_2}
      }
    }
    Retrieve the final $rr_{n}$ of the action $A_{n}$\\ \label{reward_2}
    Store \{$S_{n}, A_{n}$, $rr_{n}$, $S_{n+1}$\} and train the SAC agent\\ \label{log_data}
    \lIf{$e_n \leq \epsilon_0$}{ \label{target_perf}
        break
    }
    }   
   } 
  \textbf{Output:} Trained SAC model for online inference \label{final_model}
\end{algorithm}
\section{Simulation Setup} \label{simulation_setup}
\subsection{Network Environment Setup}
Two different types of wireless communication network environments are considered in all performed experiments, in order to showcase the robustness of our proposed approach, namely a static and a dynamic network environment. Both environments are comprised of one coordinator and up to 20 ($|\mathcal{K}| \leq 20$) heterogeneous (in terms of computational and communication capabilities) workers. 

\textbf{Static Environment Setup:}
In the static environment the $20\%$ of all workers are of reduced capabilities and are considered as low-end devices, while the rest are considered as high-end. Intuitively, the lower the resource capabilities of a worker the lower the carbon footprint and vice versa. Each low-end device $i$, operates with a maximum available computational capacity equal to  $f_i^{max} = 1$ GHz, with a total number of FLOPs per cycle equal to $c_i = 4$ \cite{9475121Mo} and a maximum transmission power equal to $p_i^{max} = 28$ dBm. In the same notion, the available computational capacity of each high-end device $j$ is up to  $f_j^{max} = 3$ GHz with a total number of FLOPs per cycle equal to $c_j = 2$ and a transmission power up to $p_j^{max}=33$ dBm. The effective switched capacitance is fixed for all workers and equal to $\varsigma_k = 10^{-28}$ Watt/Hz$\mathrm{^{3}}$ \cite{Yang13037554}. We consider non mobile workers and as such all channel gains can be considered as constant. The channel gain is modeled as $g_{k,n} = 127 + 30log_{10}(d_{k,n})$, where $d_{k,n}$ is the distance of worker $k$ from the coordinator, with white Gaussian noise power spectral density $N_0= -158$ dBm/Hz\cite{9462445Zhou}. The distance $d_k$ of each worker is fixed and randomly selected in the range of $[10,500]$ m. All workers are assigned with a fixed bandwidth that is equal to $b_{k,n} = 20$ MHz, $\forall{k} \in \mathcal{K}$ and $\forall{n}$. 

\textbf{Dynamic Environment Setup:} In the dynamic environment, in each RL episode, the percentage of low-end devices varies and follows a truncated normal distribution \cite{Robert1995} in the range $[0,60]$\% $\mathcal{N}\pm(\mu,\mu^{-},\mu^{+},\sigma^{2})$, where $(\mu,\mu^{-},\mu^{+},\sigma)=(15, -12.5, 37.5, 12)$. Due to resource contention, the availability of computational resources at workers for the FL training task can vary over time \cite{bonawitz2019}. The maximum available computational and communication capacity of the low-end devices is uniformly selected in the range $f_{i,n}^{max} \in [1,3]$ GHz and $p_{i,n}^{max} \in [23,28]$ dBm. In the same notion, the resources of the high-end devices are selected in the range $f_{j,n}^{max} \in [3.2,5]$ GHz, and $p_{j,n}^{max} \in [29,33]$ dBm, respectively. Mobile workers are also considered in this environment, where the $d_{k,n}$ of each worker dynamically changes in the range $[10,500]$ m. The allocated bandwidth of each worker also dynamically and uniformly changes in the range $b_{k,n} \in [5,20]$ MHz. 
\subsection{Federated Learning Setup}
The FL process considers two Convolutional Neural Networks (CNNs) for two benchmark datasets, namely the MNIST \cite{lecun2010mnist} and the CIFAR-10 \cite{Krizhevsky2012}. Both datasets contain 60,000 samples in total and each worker, in each RL episode, is assigned with randomly selected $\mathcal{D}_{k,n}$ non-Independent and Identically Distributed (non-IID) dataset, with $s_{k,n} \in [800,1200]$. The training is performed using the Adam optimizer, with learning rate $5\cdot10^{-5}$ and batch size 32. The ReLU is applied as activation function to all CNN and Fully Connected (FC) hidden layers. The accuracy is selected as objective function in Equations (\ref{local_accuracy_inequality}) and (\ref{global_accuracy_inequality}) on both models, with $F^* = 1$.
The configuration parameters of both CNNs are summarized in Table \ref{cnn_model}.\\ 
\textbf{MNIST FL-CNN:} The architecture consists of $658,922$ trainable parameters, with size $m = 2.51$ MB and the complexity of the model is equal to $\alpha = 1.8$ MFLOPs \cite{keras_flops}. We set $\eta = 0.5$ and $\epsilon_0 = 0.04$. The time threshold is set to $\mathsf{H} = 13$ sec. \\
\textbf{CIFAR-10 FL-CNN:} The architecture consists of $551,466$ trainable parameters, with size $m = 17.6$ MB and the complexity of the model is equal to $\alpha = 78.2$ MFLOPs. We set $\eta = 0.6$ and $\epsilon_0 = 0.04$. The time threshold is set to $\mathsf{H} = 950$ sec. 
\vspace{-4pt}

\begin{table}[ht]
\centering
\resizebox{\columnwidth}{!}{%
\begin{tabular}{|l|l|l|}
\hline
\multicolumn{1}{|c|}{\textbf{Parameter}} & \multicolumn{1}{c|}{\textbf{\begin{tabular}[c]{@{}c@{}}Configuration\\ MNIST FL-CNN\end{tabular}}} & \multicolumn{1}{c|}{\textbf{\begin{tabular}[c]{@{}c@{}}Configuration\\ CIFAR-10 FL-CNN\end{tabular}}} \\ \hline
CNN/FC Layers \textbf{x} Neurons & 1/2\textbf{x}256 & 6/1\textbf{x}128 \\ \hline
Dropout Layers & \{0.25, 0.5\} & \{0.25, 0.25,0.25,0.25\} \\ \hline
Kernel Size - Filters & (5x5) - \{16\} & (3x3) - \{32,32,64,64,128,128\} \\ \hline
Pooling Window & Max Pooling (2,2) & Max Pooling (2,2) \\ \hline
\end{tabular}%
}
\caption{FL-CNN Configuration Parameters}
\label{cnn_model}
  \vspace{-17pt}
\end{table}

\subsection{Soft Actor Critic Setup}

\begin{table}[ht]
\renewcommand{\arraystretch}{1.05}
\centering
\begin{tabular}{|c|c|} 
\hline
\textbf{Architecture} & \textbf{Configuration} \\ [0.7ex]
\hline
 Model/Policy & MLP/Stochastic\\
\hline
Policy/Value/Target DNNs & 1/2/3\\
\hline
Hidden layers/Neurons \{5,10\}-\{20\} workers & \{2/256\}-\{3/512\}\\
\hline
Batch Size/Optimizer/\textit{lr}& 256/Adam/0.001 \\
\hline
Lr Scheduler: Step Decay/Drop rate & 6000 episodes/1$\%$  \\
\hline
Entropy Coefficient & auto\_0.8 \\
\hline
Training/SDE Sample Frequency & 1000/100 RL steps \\
\hline
Training Starts & 100 RL steps \\
\hline
Replay Buffer Size & $2\cdot10^{6}$ \\
\hline
\end{tabular}
\caption{SAC Configuration}
\label{sac_model}
\vspace{-20pt}
\end{table}

The proposed SAC DRL solution is based on a custom environment interfaced through Stable-Baselines3 that provides open-source implementations of DRL algorithms in Python \cite{Raffin3546526}. The hyperparameters \cite{sac_baselines3} of the SAC DRL algorithm have been tuned, based on experimentation and are summarized in Table \ref{sac_model}. More specifically, the SAC architecture comprises one Policy, two Value and three Target Multi-Layer Perceptron (MLP) Networks, and optimizes a stochastic policy. The same MLP architecture is used for all six networks of SAC, consisting of 2 hidden layers of 256 neurons in the case of 5 and 10 workers and 3 hidden layers of 512 neurons in the case of 20 workers. The optimizer selected for the training phase of SAC is the Adam with initial value of learning rate (\textit{lr}) set to 0.001 (same for all networks) and the batch size is set to 256. A step decay scheduler is used to reduce the \textit{lr} every 6000 RL episodes by 1\%. The entropy regularization coefficient that controls the exploration/exploitation trade-off of the SAC agent is set to 0.8 and it is automatically reconfigured ('auto\_0.8') by the Stable-Baselines3 framework. The SAC agent uses State Dependent Exploration (SDE), with sample frequency equal to 100 RL steps. All SAC networks are trained every 1000 RL steps and the training starts after the first 100 RL steps, in order to fill the replay buffer of size $2\cdot10^{6}$ with enough samples. Finally, after thorough experimentation, the weight hyperparameters \{$\mu_1$, $\mu_2$\} of Eq. (\ref{penalty}) have been set to \{0.1, 0.9\}, \{0.2, 0.8\}, and \{0.4, 0.6\} for the case of 5, 10 and 20 workers, respectively.

\begin{figure*}[ht]
\centering
    \begin{subfigure}[b]{0.33\textwidth}
      \centering
      \includegraphics[width=0.91\linewidth]{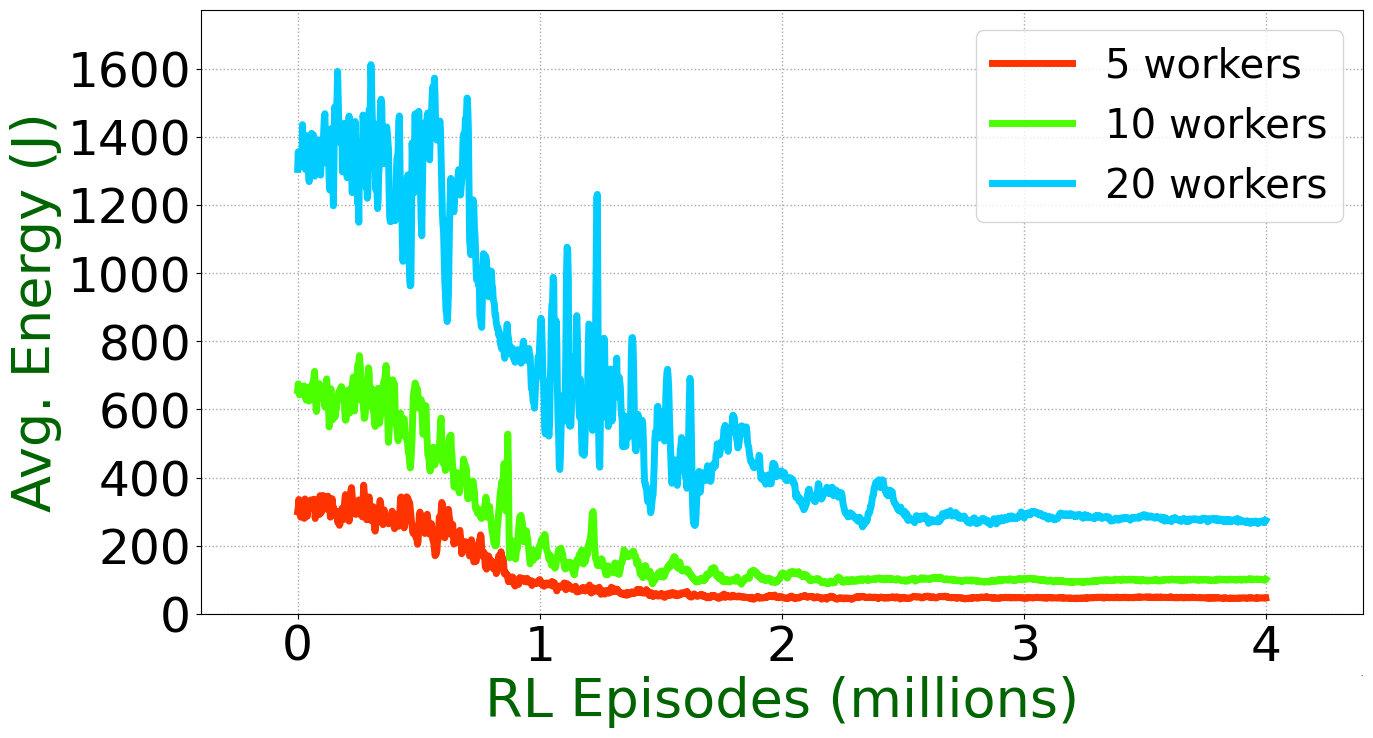}
      \caption{Average Total Energy Consumption}
      \label{fig:avg_total_energy}
    \end{subfigure}\hfill
    \begin{subfigure}[b]{0.33\textwidth}
      \centering
      \includegraphics[width=0.91\linewidth]{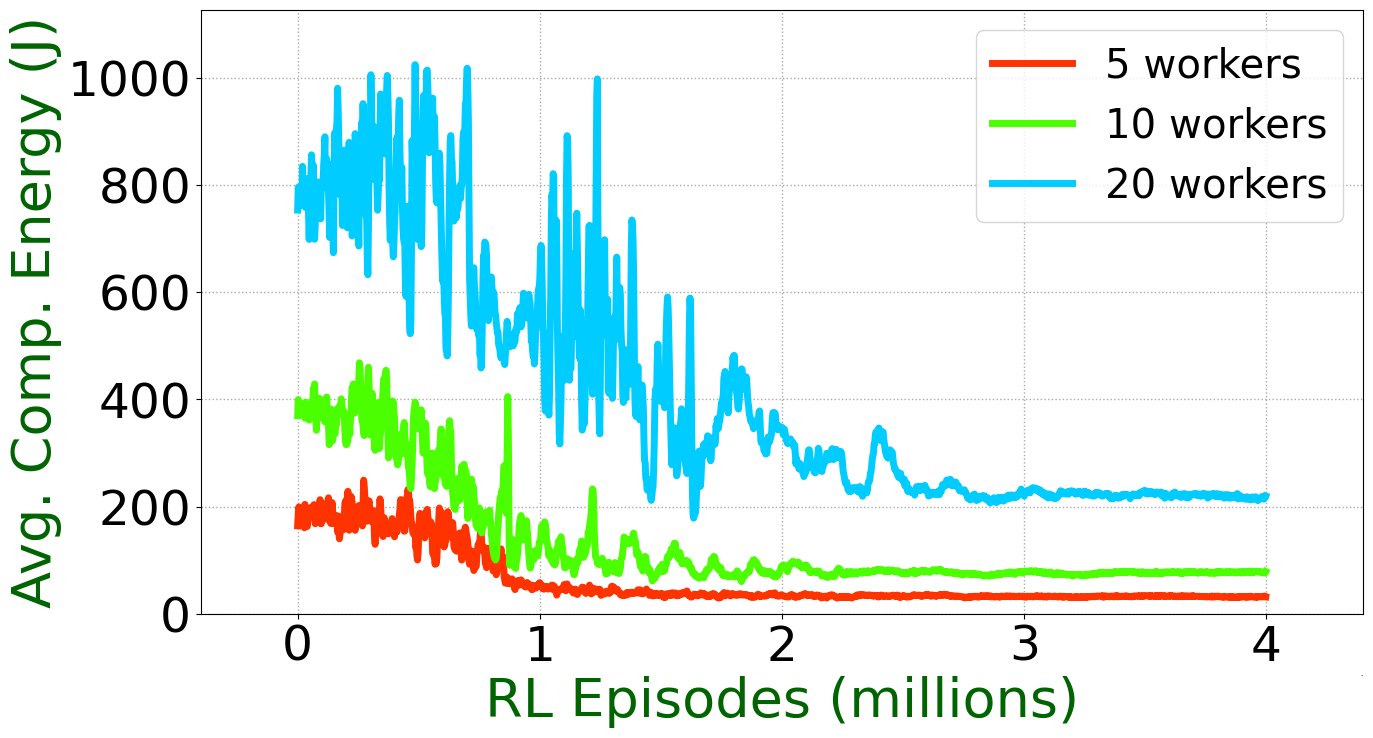}
      \caption{Average Total Computation Energy}
      \label{fig:avg_comp}
    \end{subfigure}\hfill
    \begin{subfigure}[b]{0.33\textwidth}
      \centering
      \includegraphics[width=0.91\linewidth]{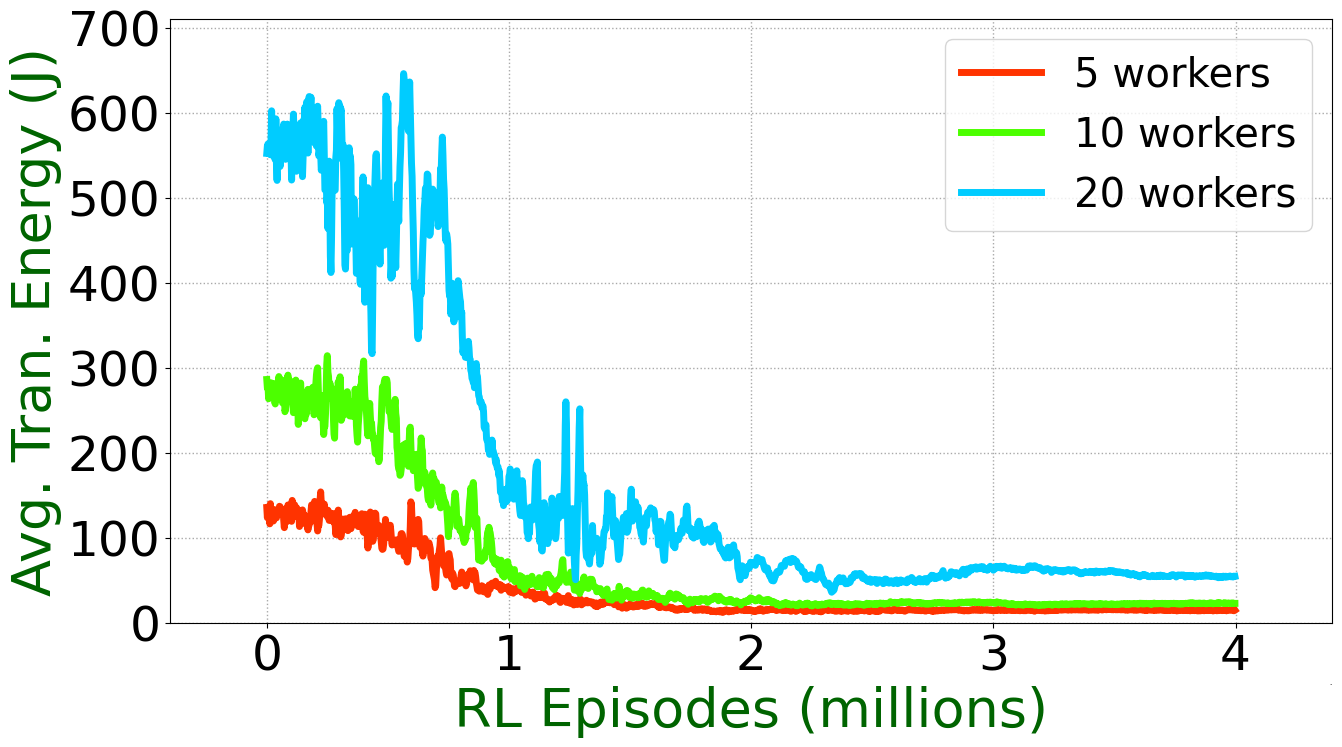}
      \caption{Average Total Transmission Energy}
      \label{fig:avg_transm}
    \end{subfigure}\hfill
    \begin{subfigure}[b]{0.33\textwidth}
      \centering
      \includegraphics[width=0.91\linewidth]{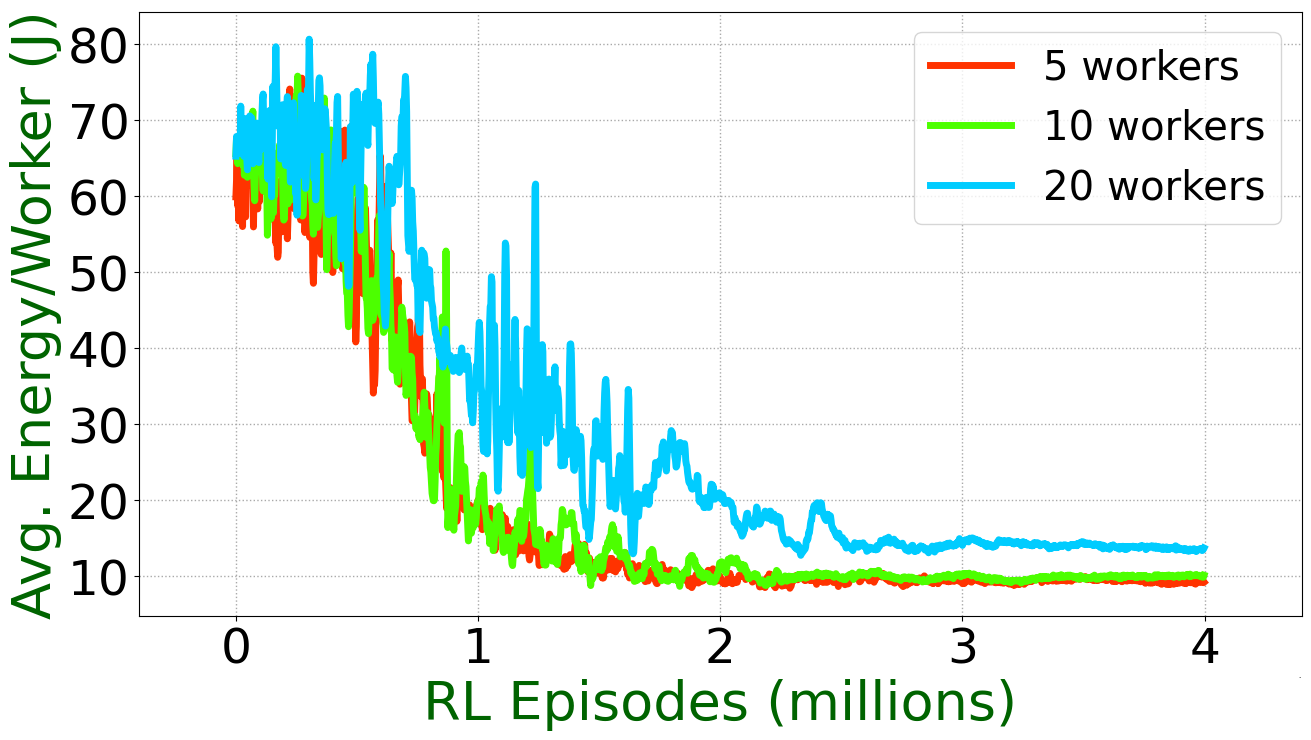}
      \caption{Average Energy Consumption per Worker}
      \label{fig:avg_per_worker_energy}
    \end{subfigure}\hfill
    \begin{subfigure}[b]{0.33\textwidth}
      \centering
      \includegraphics[width=0.91\linewidth]{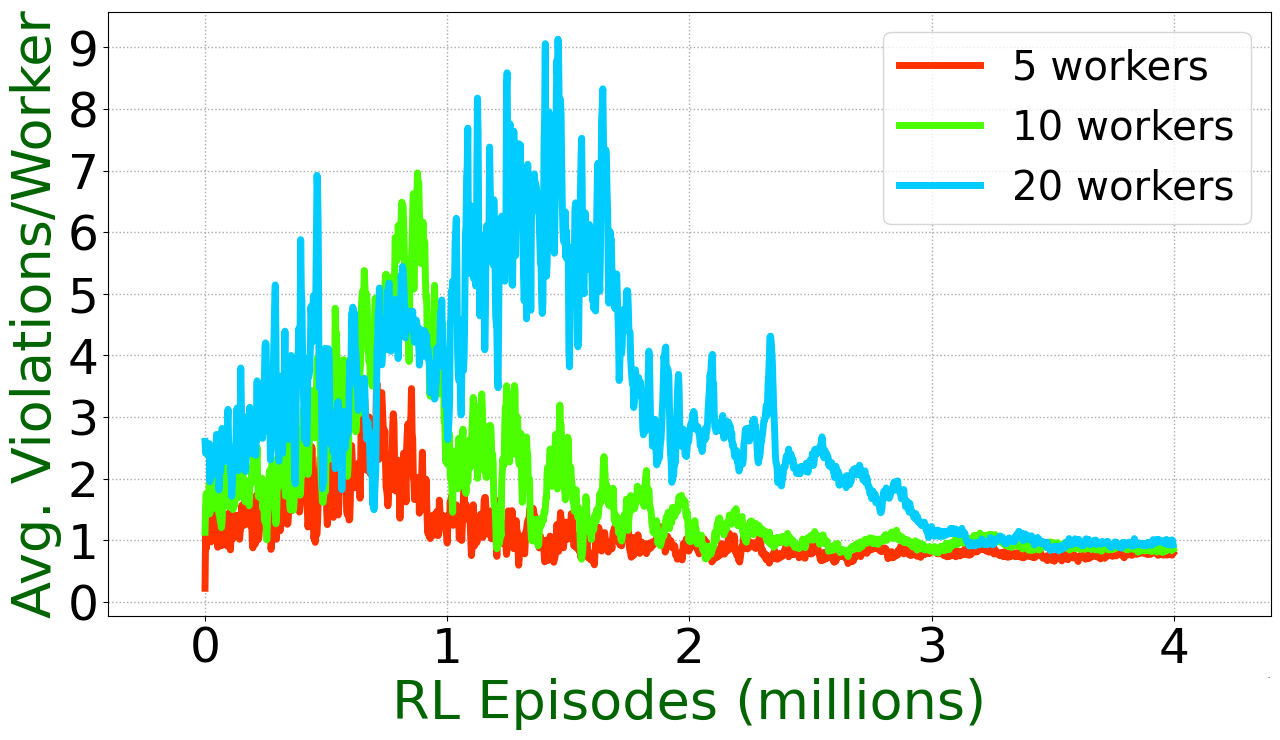}
      \caption{Average Number of Violations per Worker }
      \label{fig:avg_violations}
    \end{subfigure}\hfill
    \begin{subfigure}[b]{0.33\textwidth}
      \centering
      \includegraphics[width=0.91\linewidth]{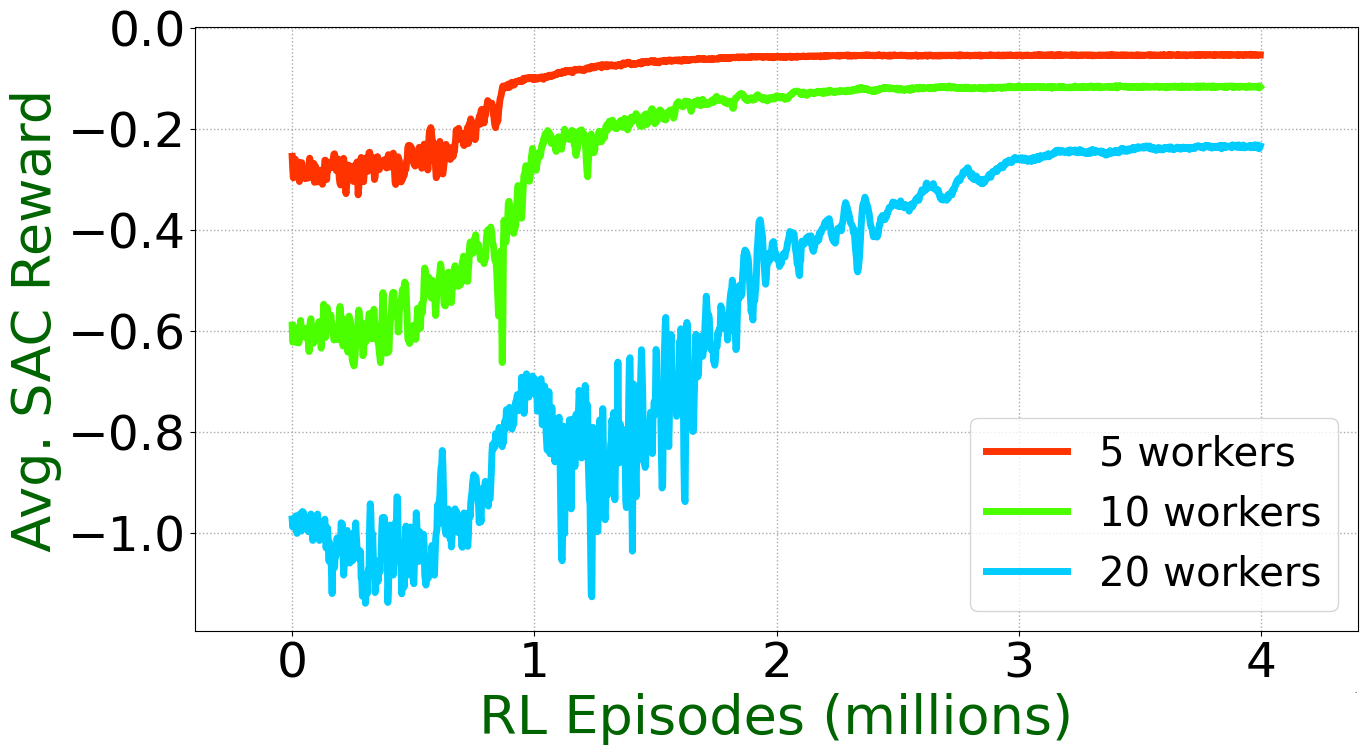}
      \caption{Average SAC Reward}
      \label{fig:avg_sac_reward}
    \end{subfigure}\hfill
    \caption{Training phase (Offline) of SAC for different number of workers (5, 10, 20)}\label{fig:train_phase}
    \vspace{-17pt}
\end{figure*}
\vspace{-5pt}
\subsection{Simulated FL Environment}\label{simulated_FL}
As already mentioned in Section \ref{proposed_solution}, for the training phase of the SAC agent, a simulated FL environment was used. In order to  produce a simulated FL environment, based on a real FL process, three main FL and energy-related parameters should be estimated. From Eq. (\ref{objective_function}), it becomes apparent that the FL parameters that are not known a priori, are the number of \textit{local} iterations per worker ($I_{k,n}$) and the total number of \textit{global} iterations ($I_0$). The simulated FL environment is based on the CNN using the MNIST dataset, since the CNN on CIFAR-10 will be used during the online inference phase to evaluate the robustness and generalization capabilities of our solution on a more complex and unknown NN architecture. Following a statistical analysis of over 100 independent FL executions, the $I_{k,n}$ is selected based on the worker's dataset variance and is in the range $[2, 11]$. Higher dataset variance, means increased data heterogeneity and as a result a higher number of \textit{local} iterations is required to reach $\eta$. According to the statistical analysis, the number of \textit{global} iterations is in the range $[10, 22]$, and it is inversely proportional to the number of \textit{local} iterations. However, since the state of the environment includes the global performance rate ($e_n$), such parameter should also be estimated. Based on Eq. (\ref{local_accuracy_inequality}), (\ref{global_accuracy_inequality}) and (\ref{global_loss}) and given the local performance target $\eta$, the estimation of the initial local accuracy of each worker ($F_k(\textbf{w}_{k,n}^{(0)},\mathcal{D}_{k,n})$) suffices to estimate the $e_n$. Once again, based on the performed statistical analysis, the initial local accuracy of each worker at the first \textit{global} iteration is randomly selected in the range $[15, 18] \%$ and it is updated appropriately at each global iteration until convergence. 
\vspace{-4pt}
\section{Performance Evaluation}\label{performance_evaluation}
\vspace{-3pt}
This section provides numerical results to evaluate the
performance and robustness of the proposed joint computation and communication
safe DRL solution under different network conditions and FL architectures. Four main performance metrics are considered in the evaluation phase: \textbf{1)} the total energy consumption of the FL (both computation and communication), \textbf{2)} the amount of wasted energy, \textbf{3)} the convergence speed of the FL (total training time) and  \textbf{4)} the channel occupancy, in terms of number of model transmissions and channel access time. 

\subsection{Offline Training Phase}
The first set of the evaluation results concerns the offline training phase of the SAC agent in the dynamic wireless network environment, using the simulated FL environment (Section \ref{simulated_FL}). 

Fig. \ref{fig:train_phase} illustrates the performance of the SAC agent, for different number of workers ($|\mathcal{K}| = \{5, 10, 20\}$)
, as the number of RL episodes increases. For visual clarity all produced results are averaged every 2500 RL episodes. Fig. \ref{fig:avg_total_energy} depicts the average total energy consumption of the FL system, that is the sum of the total computation (Fig. \ref{fig:avg_comp}) and transmission energies (Fig. \ref{fig:avg_transm}). As it can be inferred from Fig. (\ref{fig:avg_total_energy} - \ref{fig:avg_transm}), the amount of consumed energy, for the different number of workers, significantly decreases with the number of RL episodes. More specifically, the SAC agent converges to a stable solution with low energy consumption, achieving an overall reduction up to 84.6\%, 84.5\% and 79\% in the case of 5, 10, and 20 workers, respectively. As the number of workers increases, so does the time (in RL episodes) the SAC agent requires to converge. This relies on the fact that the state and action space of the environment increases proportionally to the number of workers, and as a result, the SAC agent requires more exploration time. 
The duration of the SAC agent's high exploration phase lasts approximately 0.9, 2.1 and 3.2 million episodes, in the case of 5, 10 and 20 workers, respectively. 

In addition, it is intuitive that the higher the number of workers, the higher the total energy consumption of the system. As a result, the average reward that the SAC agent retrieves from the environment is lower for higher number of workers (Fig. \ref{fig:avg_sac_reward}). However, as it is depicted in Fig. \ref{fig:avg_per_worker_energy}, the SAC agent achieves an average amount of total energy consumption per worker approximately equal to 11 Joules, in all three cases. This behavior showcases the scalability of our solution, since the increase in the total amount of energy consumption is mainly based on the number of workers. In addition, the slight variations are explained by the fact that the number of \textit{global} iterations increases with the number of workers. More specifically, the average number of \textit{global} iterations in the case of 5, 10 and 20 workers is equal to 15, 17 and 19 with a standard deviation of 4, 2 and 2, respectively. This increase in the number of \textit{global} iterations is experienced due to the non-IID dataset that can affect the convergence speed of the FL process \cite{MA2022244}. 

As already stated, the current work introduced a penalty function towards a safe RL process. Based on this, Fig. \ref{fig:avg_violations} depicts the average number of constraint violations per worker in an FL process. This figure validates our claims towards a better decision making for the SAC agent, since the total number of constraint violations of each worker during an FL process is almost minimized, achieving on average 0.79, 0.85 and 0.92 violations per worker in the case of 5, 10 and 20 workers, respectively. Despite the dynamicity of the network environment the SAC RL agent manages to converge to a safe policy that enables an efficient FL process, where all workers transmit their weight updates on time without wasting resources, while collaboratively respecting the imposed training performance target of the global FL model $\epsilon0$.

To summarize, during the offline training phase, the SAC agent showcases a scalable performance in a dynamic training environment, retaining a similar energy consumption per worker for all three cases equal to 11 Joules. Finally, the proposed safe DRL approach highly contributes to an almost violations-free FL process, resulting on average in 0.85 constraint violations per worker for all three cases.
\vspace{-3pt}
\subsection{Online Inference Phase}
\vspace{-2pt}
\begin{table*}[h]
\renewcommand{\arraystretch}{1}
\centering
\resizebox{\linewidth}{!}{%
\begin{tabular}{|
>{\columncolor[HTML]{FFFFFF}}l |
>{\columncolor[HTML]{EFEFEF}}c 
>{\columncolor[HTML]{FFFFFF}}c 
>{\columncolor[HTML]{FFFFFF}}c 
>{\columncolor[HTML]{FFFFFF}}c 
>{\columncolor[HTML]{FFFFFF}}c |
>{\columncolor[HTML]{EFEFEF}}c 
>{\columncolor[HTML]{FFFFFF}}c 
>{\columncolor[HTML]{FFFFFF}}c 
>{\columncolor[HTML]{FFFFFF}}c 
>{\columncolor[HTML]{FFFFFF}}c |
>{\columncolor[HTML]{EFEFEF}}c 
>{\columncolor[HTML]{FFFFFF}}c 
>{\columncolor[HTML]{FFFFFF}}c 
>{\columncolor[HTML]{FFFFFF}}c 
>{\columncolor[HTML]{FFFFFF}}c |}
\hline
\multicolumn{1}{|c|}{\cellcolor[HTML]{FFFFFF}\textbf{\begin{tabular}[c]{@{}c@{}}Total Avg. \\ ($\pm$STD)\end{tabular}}} &
  \multicolumn{5}{c|}{\cellcolor[HTML]{FFFFFF}\textbf{5 Workers}} &
  \multicolumn{5}{c|}{\cellcolor[HTML]{FFFFFF}\textbf{10 Workers}} &
  \multicolumn{5}{c|}{\cellcolor[HTML]{FFFFFF}\textbf{20 Workers}} \\ \hline
\textbf{Schedulers} &
  \multicolumn{1}{c|}{\cellcolor[HTML]{EFEFEF}\textbf{SAC}} &
  \multicolumn{1}{c|}{\cellcolor[HTML]{FFFFFF}\textbf{GAS}} &
  \multicolumn{1}{c|}{\cellcolor[HTML]{FFFFFF}\textbf{BES}} &
  \multicolumn{1}{c|}{\cellcolor[HTML]{FFFFFF}\textbf{RSS}} &
  \textbf{GSS} &
  \multicolumn{1}{c|}{\cellcolor[HTML]{EFEFEF}\textbf{SAC}} &
  \multicolumn{1}{c|}{\cellcolor[HTML]{FFFFFF}\textbf{GAS}} &
  \multicolumn{1}{c|}{\cellcolor[HTML]{FFFFFF}\textbf{BES}} &
  \multicolumn{1}{c|}{\cellcolor[HTML]{FFFFFF}\textbf{RSS}} &
  \textbf{GSS} &
  \multicolumn{1}{c|}{\cellcolor[HTML]{EFEFEF}\textbf{SAC}} &
  \multicolumn{1}{c|}{\cellcolor[HTML]{FFFFFF}\textbf{GAS}} &
  \multicolumn{1}{c|}{\cellcolor[HTML]{FFFFFF}\textbf{BES}} &
  \multicolumn{1}{c|}{\cellcolor[HTML]{FFFFFF}\textbf{RSS}} &
  \textbf{GSS} \\ \hline
  \textbf{\begin{tabular}[c]{@{}l@{}}Total\\ Energy (J)\end{tabular}} &
  \multicolumn{1}{c|}{\cellcolor[HTML]{EFEFEF}\begin{tabular}[c]{@{}c@{}}15.8\\ ($\pm$5.6)\end{tabular}} &
  \multicolumn{1}{c|}{\cellcolor[HTML]{FFFFFF}\begin{tabular}[c]{@{}c@{}}19.3\\ ($\pm$4.5)\end{tabular}} &
  \multicolumn{1}{c|}{\cellcolor[HTML]{FFFFFF}\begin{tabular}[c]{@{}c@{}}292.7\\ ($\pm$69.5)\end{tabular}} &
  \multicolumn{1}{c|}{\cellcolor[HTML]{FFFFFF}\begin{tabular}[c]{@{}c@{}}116.4\\ ($\pm$23.2)\end{tabular}} &
  \begin{tabular}[c]{@{}c@{}}51.5\\ ($\pm$14.6)\end{tabular} &
  \multicolumn{1}{c|}{\cellcolor[HTML]{EFEFEF}\begin{tabular}[c]{@{}c@{}}35.7\\ ($\pm$7.9)\end{tabular}} &
  \multicolumn{1}{c|}{\cellcolor[HTML]{FFFFFF}\begin{tabular}[c]{@{}c@{}}52.8\\ ($\pm$7\end{tabular}} &
  \multicolumn{1}{c|}{\cellcolor[HTML]{FFFFFF}\begin{tabular}[c]{@{}c@{}}664.4\\ ($\pm$91.9)\end{tabular}} &
  \multicolumn{1}{c|}{\cellcolor[HTML]{FFFFFF}\begin{tabular}[c]{@{}c@{}}271.1\\ ($\pm$38)\end{tabular}} &
  \begin{tabular}[c]{@{}c@{}}168\\ ($\pm$27.9)\end{tabular} &
  \multicolumn{1}{c|}{\cellcolor[HTML]{EFEFEF}\begin{tabular}[c]{@{}c@{}}97.9\\ ($\pm$15.5)\end{tabular}} &
  \multicolumn{1}{c|}{\cellcolor[HTML]{FFFFFF}\begin{tabular}[c]{@{}c@{}}136.7\\ ($\pm$12.9)\end{tabular}} &
  \multicolumn{1}{c|}{\cellcolor[HTML]{FFFFFF}\begin{tabular}[c]{@{}c@{}}1407.2\\ ($\pm$144.7)\end{tabular}} &
  \multicolumn{1}{c|}{\cellcolor[HTML]{FFFFFF}\begin{tabular}[c]{@{}c@{}}576.4\\ ($\pm$69.4)\end{tabular}} &
  \begin{tabular}[c]{@{}c@{}}400.7\\ ($\pm$52.3)\end{tabular} \\ \hline
\textbf{\begin{tabular}[c]{@{}l@{}}Computation\\ Energy (J)\end{tabular}} &
  \multicolumn{1}{c|}{\cellcolor[HTML]{EFEFEF}\begin{tabular}[c]{@{}c@{}}8.7\\ ($\pm$3.8)\end{tabular}} &
  \multicolumn{1}{c|}{\cellcolor[HTML]{FFFFFF}\begin{tabular}[c]{@{}c@{}}7.5\\ ($\pm$1.7)\end{tabular}} &
  \multicolumn{1}{c|}{\cellcolor[HTML]{FFFFFF}\begin{tabular}[c]{@{}c@{}}153\\ ($\pm$38.3)\end{tabular}} &
  \multicolumn{1}{c|}{\cellcolor[HTML]{FFFFFF}\begin{tabular}[c]{@{}c@{}}51.1\\ ($\pm$12)\end{tabular}} &
  \begin{tabular}[c]{@{}c@{}}25.3\\ ($\pm$7.5)\end{tabular} &
  \multicolumn{1}{c|}{\cellcolor[HTML]{EFEFEF}\begin{tabular}[c]{@{}c@{}}23.7\\ ($\pm$6.5)\end{tabular}} &
  \multicolumn{1}{c|}{\cellcolor[HTML]{FFFFFF}\begin{tabular}[c]{@{}c@{}}30.5\\ ($\pm$4.4\end{tabular}} &
  \multicolumn{1}{c|}{\cellcolor[HTML]{FFFFFF}\begin{tabular}[c]{@{}c@{}}348\\ ($\pm$51.2)\end{tabular}} &
  \multicolumn{1}{c|}{\cellcolor[HTML]{FFFFFF}\begin{tabular}[c]{@{}c@{}}120.6\\ ($\pm$19.7)\end{tabular}} &
  \begin{tabular}[c]{@{}c@{}}112.2\\ ($\pm$20)\end{tabular} &
  \multicolumn{1}{c|}{\cellcolor[HTML]{EFEFEF}\begin{tabular}[c]{@{}c@{}}65.9\\ ($\pm$13.1)\end{tabular}} &
  \multicolumn{1}{c|}{\cellcolor[HTML]{FFFFFF}\begin{tabular}[c]{@{}c@{}}61.4\\ ($\pm$6.5)\end{tabular}} &
  \multicolumn{1}{c|}{\cellcolor[HTML]{FFFFFF}\begin{tabular}[c]{@{}c@{}}757\\ ($\pm$89.4)\end{tabular}} &
  \multicolumn{1}{c|}{\cellcolor[HTML]{FFFFFF}\begin{tabular}[c]{@{}c@{}}264.4\\ ($\pm$32.8)\end{tabular}} &
  \begin{tabular}[c]{@{}c@{}}194.2\\ ($\pm$27.8)\end{tabular} \\ \hline
\textbf{\begin{tabular}[c]{@{}l@{}}Transmission\\ Energy (J)\end{tabular}} &
  \multicolumn{1}{c|}{\cellcolor[HTML]{EFEFEF}\begin{tabular}[c]{@{}c@{}}7.2\\ ($\pm$3.1)\end{tabular}} &
  \multicolumn{1}{c|}{\cellcolor[HTML]{FFFFFF}\begin{tabular}[c]{@{}c@{}}11.8\\ ($\pm$3.2)\end{tabular}} &
  \multicolumn{1}{c|}{\cellcolor[HTML]{FFFFFF}\begin{tabular}[c]{@{}c@{}}139.6\\ ($\pm$37.5)\end{tabular}} &
  \multicolumn{1}{c|}{\cellcolor[HTML]{FFFFFF}\begin{tabular}[c]{@{}c@{}}65.3\\ ($\pm$15.4)\end{tabular}} &
  \begin{tabular}[c]{@{}c@{}}26.2\\ ($\pm$8.4)\end{tabular} &
  \multicolumn{1}{c|}{\cellcolor[HTML]{EFEFEF}\begin{tabular}[c]{@{}c@{}}12\\ ($\pm$3.4)\end{tabular}} &
  \multicolumn{1}{c|}{\cellcolor[HTML]{FFFFFF}\begin{tabular}[c]{@{}c@{}}22.3\\ ($\pm$3.3)\end{tabular}} &
  \multicolumn{1}{c|}{\cellcolor[HTML]{FFFFFF}\begin{tabular}[c]{@{}c@{}}316.4\\ ($\pm$50.2)\end{tabular}} &
  \multicolumn{1}{c|}{\cellcolor[HTML]{FFFFFF}\begin{tabular}[c]{@{}c@{}}150.5\\ ($\pm$24.1)\end{tabular}} &
  \begin{tabular}[c]{@{}c@{}}55.8\\ ($\pm$12.3)\end{tabular} &
  \multicolumn{1}{c|}{\cellcolor[HTML]{EFEFEF}\begin{tabular}[c]{@{}c@{}}32\\ ($\pm$6)\end{tabular}} &
  \multicolumn{1}{c|}{\cellcolor[HTML]{FFFFFF}\begin{tabular}[c]{@{}c@{}}75.4\\ ($\pm$8.4)\end{tabular}} &
  \multicolumn{1}{c|}{\cellcolor[HTML]{FFFFFF}\begin{tabular}[c]{@{}c@{}}650.2\\ ($\pm$73.7)\end{tabular}} &
  \multicolumn{1}{c|}{\cellcolor[HTML]{FFFFFF}\begin{tabular}[c]{@{}c@{}}312\\ ($\pm$43.9)\end{tabular}} &
  \begin{tabular}[c]{@{}c@{}}206.5\\ ($\pm$31.9)\end{tabular} \\ \hline
\textbf{\begin{tabular}[c]{@{}l@{}}Training Time\\ per Global \\ Iteration (s)\end{tabular}} &
  \multicolumn{1}{c|}{\cellcolor[HTML]{EFEFEF}\begin{tabular}[c]{@{}c@{}}9.7\\ ($\pm$2.3)\end{tabular}} &
  \multicolumn{1}{c|}{\cellcolor[HTML]{FFFFFF}\begin{tabular}[c]{@{}c@{}}8.1\\ ($\pm$2.5)\end{tabular}} &
  \multicolumn{1}{c|}{\cellcolor[HTML]{FFFFFF}\begin{tabular}[c]{@{}c@{}}2.7\\ ($\pm$0.6)\end{tabular}} &
  \multicolumn{1}{c|}{\cellcolor[HTML]{FFFFFF}\begin{tabular}[c]{@{}c@{}}9\\ ($\pm$1.2)\end{tabular}} &
  \begin{tabular}[c]{@{}c@{}}8.9\\ ($\pm$2.1)\end{tabular} &
  \multicolumn{1}{c|}{\cellcolor[HTML]{EFEFEF}\begin{tabular}[c]{@{}c@{}}10.1\\ ($\pm$1.8)\end{tabular}} &
  \multicolumn{1}{c|}{\cellcolor[HTML]{FFFFFF}\begin{tabular}[c]{@{}c@{}}7.2\\ ($\pm$1.7)\end{tabular}} &
  \multicolumn{1}{c|}{\cellcolor[HTML]{FFFFFF}\begin{tabular}[c]{@{}c@{}}3\\ ($\pm$0.4)\end{tabular}} &
  \multicolumn{1}{c|}{\cellcolor[HTML]{FFFFFF}\begin{tabular}[c]{@{}c@{}}11\\ ($\pm$0.8)\end{tabular}} &
  \begin{tabular}[c]{@{}c@{}}7.6\\ ($\pm$1.5)\end{tabular} &
  \multicolumn{1}{c|}{\cellcolor[HTML]{EFEFEF}\begin{tabular}[c]{@{}c@{}}11.2\\ ($\pm$1.5)\end{tabular}} &
  \multicolumn{1}{c|}{\cellcolor[HTML]{FFFFFF}\begin{tabular}[c]{@{}c@{}}8.9\\ ($\pm$1)\end{tabular}} &
  \multicolumn{1}{c|}{\cellcolor[HTML]{FFFFFF}\begin{tabular}[c]{@{}c@{}}3.4\\ ($\pm$0.5)\end{tabular}} &
  \multicolumn{1}{c|}{\cellcolor[HTML]{FFFFFF}\begin{tabular}[c]{@{}c@{}}12.5\\ ($\pm$0.4)\end{tabular}} &
  \begin{tabular}[c]{@{}c@{}}11.9\\ ($\pm$1.1)\end{tabular} \\ \hline
\textbf{\begin{tabular}[c]{@{}l@{}}Global\\ Iterations\end{tabular}} &
  \multicolumn{1}{c|}{\cellcolor[HTML]{EFEFEF}\begin{tabular}[c]{@{}c@{}}15.7\\ ($\pm$4.4)\end{tabular}} &
  \multicolumn{1}{c|}{\cellcolor[HTML]{FFFFFF}\begin{tabular}[c]{@{}c@{}}15.4\\ ($\pm$4)\end{tabular}} &
  \multicolumn{1}{c|}{\cellcolor[HTML]{FFFFFF}\begin{tabular}[c]{@{}c@{}}15.4\\ ($\pm$4.1)\end{tabular}} &
  \multicolumn{1}{c|}{\cellcolor[HTML]{FFFFFF}\begin{tabular}[c]{@{}c@{}}15.6\\ ($\pm$3.5)\end{tabular}} &
  \begin{tabular}[c]{@{}c@{}}15.7\\ ($\pm$4.4)\end{tabular} &
  \multicolumn{1}{c|}{\cellcolor[HTML]{EFEFEF}\begin{tabular}[c]{@{}c@{}}18\\ ($\pm$2.9)\end{tabular}} &
  \multicolumn{1}{c|}{\cellcolor[HTML]{FFFFFF}\begin{tabular}[c]{@{}c@{}}17.6\\ ($\pm$2.6)\end{tabular}} &
  \multicolumn{1}{c|}{\cellcolor[HTML]{FFFFFF}\begin{tabular}[c]{@{}c@{}}17.5\\ ($\pm$2.8)\end{tabular}} &
  \multicolumn{1}{c|}{\cellcolor[HTML]{FFFFFF}\begin{tabular}[c]{@{}c@{}}18.1\\ ($\pm$2.7)\end{tabular}} &
  \begin{tabular}[c]{@{}c@{}}17.4\\ ($\pm$2.8)\end{tabular} &
  \multicolumn{1}{c|}{\cellcolor[HTML]{EFEFEF}\begin{tabular}[c]{@{}c@{}}18.6\\ ($\pm$2.3)\end{tabular}} &
  \multicolumn{1}{c|}{\cellcolor[HTML]{FFFFFF}\begin{tabular}[c]{@{}c@{}}18\\ ($\pm$2)\end{tabular}} &
  \multicolumn{1}{c|}{\cellcolor[HTML]{FFFFFF}\begin{tabular}[c]{@{}c@{}}17.9\\ ($\pm$2)\end{tabular}} &
  \multicolumn{1}{c|}{\cellcolor[HTML]{FFFFFF}\begin{tabular}[c]{@{}c@{}}18.7\\ ($\pm$2.5)\end{tabular}} &
  \begin{tabular}[c]{@{}c@{}}18.4\\ ($\pm$2.5)\end{tabular} \\ \hline
\end{tabular}%
}
\vspace{-2pt}
\caption{\textbf{Static Environment - MNIST}: Comparison of our solution (SAC) against four baseline schedulers }
\label{tab:baseline_comparison_static}
\vspace{-6pt}
\end{table*}
\begin{table*}[h]
\renewcommand{\arraystretch}{1}
\centering
\resizebox{\linewidth}{!}{%
\begin{tabular}{|l|ccccc|ccccc|ccccc|}
\hline
\multicolumn{1}{|c|}{\textbf{\begin{tabular}[c]{@{}c@{}}Total Avg. \\ ($\pm$STD)\end{tabular}}} &
  \multicolumn{5}{c|}{\textbf{5 Workers}} &
  \multicolumn{5}{c|}{\textbf{10 Workers}} &
  \multicolumn{5}{c|}{\textbf{20 Workers}} \\ \hline
\textbf{Schedulers} &
  \multicolumn{1}{c|}{\cellcolor[HTML]{EFEFEF}\textbf{SAC}} &
  \multicolumn{1}{c|}{\cellcolor[HTML]{FFFFFF}\textbf{GAS}} &
  \multicolumn{1}{c|}{\cellcolor[HTML]{FFFFFF}\textbf{BES}} &
  \multicolumn{1}{c|}{\cellcolor[HTML]{FFFFFF}\textbf{RSS}} &
  \cellcolor[HTML]{FFFFFF}\textbf{GSS} &
  \multicolumn{1}{c|}{\cellcolor[HTML]{EFEFEF}\textbf{SAC}} &
  \multicolumn{1}{c|}{\cellcolor[HTML]{FFFFFF}\textbf{GAS}} &
  \multicolumn{1}{c|}{\cellcolor[HTML]{FFFFFF}\textbf{BES}} &
  \multicolumn{1}{c|}{\cellcolor[HTML]{FFFFFF}\textbf{RSS}} &
  \cellcolor[HTML]{FFFFFF}\textbf{GSS} &
  \multicolumn{1}{c|}{\cellcolor[HTML]{EFEFEF}\textbf{SAC}} &
  \multicolumn{1}{c|}{\cellcolor[HTML]{FFFFFF}\textbf{GAS}} &
  \multicolumn{1}{c|}{\cellcolor[HTML]{FFFFFF}\textbf{BES}} &
  \multicolumn{1}{c|}{\cellcolor[HTML]{FFFFFF}\textbf{RSS}} &
  \cellcolor[HTML]{FFFFFF}\textbf{GSS} \\ \hline
  \textbf{\begin{tabular}[c]{@{}l@{}}Total\\ Energy (J)\end{tabular}} &
  \multicolumn{1}{c|}{\cellcolor[HTML]{EFEFEF}\begin{tabular}[c]{@{}c@{}}35.1\\ ($\pm$12.6)\end{tabular}} &
  \multicolumn{1}{c|}{\cellcolor[HTML]{FFFFFF}\begin{tabular}[c]{@{}c@{}}53.1\\ ($\pm$20.8)\end{tabular}} &
  \multicolumn{1}{c|}{\cellcolor[HTML]{FFFFFF}\begin{tabular}[c]{@{}c@{}}603.5\\ ($\pm$172.3)\end{tabular}} &
  \multicolumn{1}{c|}{\cellcolor[HTML]{FFFFFF}\begin{tabular}[c]{@{}c@{}}237.7\\ ($\pm$74)\end{tabular}} &
  \cellcolor[HTML]{FFFFFF}\begin{tabular}[c]{@{}c@{}}148\\ ($\pm$54.5)\end{tabular} &
  \multicolumn{1}{c|}{\cellcolor[HTML]{EFEFEF}\begin{tabular}[c]{@{}c@{}}86.6\\ ($\pm$21.7)\end{tabular}} &
  \multicolumn{1}{c|}{\cellcolor[HTML]{FFFFFF}\begin{tabular}[c]{@{}c@{}}118.7\\ ($\pm$26.1)\end{tabular}} &
  \multicolumn{1}{c|}{\cellcolor[HTML]{FFFFFF}\begin{tabular}[c]{@{}c@{}}1343.7\\ ($\pm$250)\end{tabular}} &
  \multicolumn{1}{c|}{\cellcolor[HTML]{FFFFFF}\begin{tabular}[c]{@{}c@{}}542.5\\ ($\pm$103.4)\end{tabular}} &
  \cellcolor[HTML]{FFFFFF}\begin{tabular}[c]{@{}c@{}}465\\ ($\pm$95.8)\end{tabular} &
  \multicolumn{1}{c|}{\cellcolor[HTML]{EFEFEF}\begin{tabular}[c]{@{}c@{}}195\\ ($\pm$53)\end{tabular}} &
  \multicolumn{1}{c|}{\cellcolor[HTML]{FFFFFF}\begin{tabular}[c]{@{}c@{}}355.1\\ ($\pm$100.3)\end{tabular}} &
  \multicolumn{1}{c|}{\cellcolor[HTML]{FFFFFF}\begin{tabular}[c]{@{}c@{}}2819.8\\ ($\pm$442.5)\end{tabular}} &
  \multicolumn{1}{c|}{\cellcolor[HTML]{FFFFFF}\begin{tabular}[c]{@{}c@{}}1130.8\\ ($\pm$184.7)\end{tabular}} &
  \cellcolor[HTML]{FFFFFF}\begin{tabular}[c]{@{}c@{}}932.9\\ ($\pm$157.1)\end{tabular} \\ \hline
\textbf{\begin{tabular}[c]{@{}l@{}}Computation\\ Energy (J)\end{tabular}} &
  \multicolumn{1}{c|}{\cellcolor[HTML]{EFEFEF}\begin{tabular}[c]{@{}c@{}}19.5\\ ($\pm$8.3)\end{tabular}} &
  \multicolumn{1}{c|}{\cellcolor[HTML]{FFFFFF}\begin{tabular}[c]{@{}c@{}}29.1\\ ($\pm$15.1)\end{tabular}} &
  \multicolumn{1}{c|}{\cellcolor[HTML]{FFFFFF}\begin{tabular}[c]{@{}c@{}}324.7\\ ($\pm$92.4)\end{tabular}} &
  \multicolumn{1}{c|}{\cellcolor[HTML]{FFFFFF}\begin{tabular}[c]{@{}c@{}}107.5\\ ($\pm$34.1)\end{tabular}} &
  \cellcolor[HTML]{FFFFFF}\begin{tabular}[c]{@{}c@{}}61.5\\ ($\pm$29.5)\end{tabular} &
  \multicolumn{1}{c|}{\cellcolor[HTML]{EFEFEF}\begin{tabular}[c]{@{}c@{}}59\\ ($\pm$16.6)\end{tabular}} &
  \multicolumn{1}{c|}{\cellcolor[HTML]{FFFFFF}\begin{tabular}[c]{@{}c@{}}71.8\\ ($\pm$17.5)\end{tabular}} &
  \multicolumn{1}{c|}{\cellcolor[HTML]{FFFFFF}\begin{tabular}[c]{@{}c@{}}724.1\\ ($\pm$143.6)\end{tabular}} &
  \multicolumn{1}{c|}{\cellcolor[HTML]{FFFFFF}\begin{tabular}[c]{@{}c@{}}245.2\\ ($\pm$53.6)\end{tabular}} &
  \cellcolor[HTML]{FFFFFF}\begin{tabular}[c]{@{}c@{}}285.4\\ ($\pm$67.8)\end{tabular} &
  \multicolumn{1}{c|}{\cellcolor[HTML]{EFEFEF}\begin{tabular}[c]{@{}c@{}}144.5\\ ($\pm$78.7)\end{tabular}} &
  \multicolumn{1}{c|}{\cellcolor[HTML]{FFFFFF}\begin{tabular}[c]{@{}c@{}}180.2\\ ($\pm$63.8)\end{tabular}} &
  \multicolumn{1}{c|}{\cellcolor[HTML]{FFFFFF}\begin{tabular}[c]{@{}c@{}}1553.4\\ ($\pm$283.1)\end{tabular}} &
  \multicolumn{1}{c|}{\cellcolor[HTML]{FFFFFF}\begin{tabular}[c]{@{}c@{}}527.9\\ ($\pm$91.4)\end{tabular}} &
  \cellcolor[HTML]{FFFFFF}\begin{tabular}[c]{@{}c@{}}404.8\\ ($\pm$83.4)\end{tabular} \\ \hline
\textbf{\begin{tabular}[c]{@{}l@{}}Transmission\\ Energy (J)\end{tabular}} &
  \multicolumn{1}{c|}{\cellcolor[HTML]{EFEFEF}\begin{tabular}[c]{@{}c@{}}15.6\\ ($\pm$7.8)\end{tabular}} &
  \multicolumn{1}{c|}{\cellcolor[HTML]{FFFFFF}\begin{tabular}[c]{@{}c@{}}24\\ ($\pm$8.4)\end{tabular}} &
  \multicolumn{1}{c|}{\cellcolor[HTML]{FFFFFF}\begin{tabular}[c]{@{}c@{}}278.8\\ ($\pm$100.4)\end{tabular}} &
  \multicolumn{1}{c|}{\cellcolor[HTML]{FFFFFF}\begin{tabular}[c]{@{}c@{}}130.2\\ ($\pm$49.5)\end{tabular}} &
  \cellcolor[HTML]{FFFFFF}\begin{tabular}[c]{@{}c@{}}86.5\\ ($\pm$33.9)\end{tabular} &
  \multicolumn{1}{c|}{\cellcolor[HTML]{EFEFEF}\begin{tabular}[c]{@{}c@{}}27.6\\ ($\pm$9.7)\end{tabular}} &
  \multicolumn{1}{c|}{\cellcolor[HTML]{FFFFFF}\begin{tabular}[c]{@{}c@{}}46.9\\ ($\pm$11.4)\end{tabular}} &
  \multicolumn{1}{c|}{\cellcolor[HTML]{FFFFFF}\begin{tabular}[c]{@{}c@{}}619.6\\ ($\pm$142)\end{tabular}} &
  \multicolumn{1}{c|}{\cellcolor[HTML]{FFFFFF}\begin{tabular}[c]{@{}c@{}}297.3\\ ($\pm$68.1)\end{tabular}} &
  \cellcolor[HTML]{FFFFFF}\begin{tabular}[c]{@{}c@{}}179.6\\ ($\pm$62.2)\end{tabular} &
  \multicolumn{1}{c|}{\cellcolor[HTML]{EFEFEF}\begin{tabular}[c]{@{}c@{}}50.4\\ ($\pm$11.1)\end{tabular}} &
  \multicolumn{1}{c|}{\cellcolor[HTML]{FFFFFF}\begin{tabular}[c]{@{}c@{}}174.9\\ ($\pm$46.1)\end{tabular}} &
  \multicolumn{1}{c|}{\cellcolor[HTML]{FFFFFF}\begin{tabular}[c]{@{}c@{}}1266.4\\ ($\pm$209.3)\end{tabular}} &
  \multicolumn{1}{c|}{\cellcolor[HTML]{FFFFFF}\begin{tabular}[c]{@{}c@{}}602.9\\ ($\pm$114.8)\end{tabular}} &
  \cellcolor[HTML]{FFFFFF}\begin{tabular}[c]{@{}c@{}}528.1\\ ($\pm$97)\end{tabular} \\ \hline
\textbf{\begin{tabular}[c]{@{}l@{}}Training Time\\ per Global \\ Iteration (s)\end{tabular}} &
  \multicolumn{1}{c|}{\cellcolor[HTML]{EFEFEF}\begin{tabular}[c]{@{}c@{}}8.5\\ ($\pm$2.1)\end{tabular}} &
  \multicolumn{1}{c|}{\cellcolor[HTML]{FFFFFF}\begin{tabular}[c]{@{}c@{}}7.6\\ ($\pm$1.6)\end{tabular}} &
  \multicolumn{1}{c|}{\cellcolor[HTML]{FFFFFF}\begin{tabular}[c]{@{}c@{}}3.7\\ ($\pm$0.7)\end{tabular}} &
  \multicolumn{1}{c|}{\cellcolor[HTML]{FFFFFF}\begin{tabular}[c]{@{}c@{}}8.3\\ ($\pm$1.1)\end{tabular}} &
  \cellcolor[HTML]{FFFFFF}\begin{tabular}[c]{@{}c@{}}8.2\\ ($\pm$2.4)\end{tabular} &
  \multicolumn{1}{c|}{\cellcolor[HTML]{EFEFEF}\begin{tabular}[c]{@{}c@{}}8.4\\ ($\pm$1.7)\end{tabular}} &
  \multicolumn{1}{c|}{\cellcolor[HTML]{FFFFFF}\begin{tabular}[c]{@{}c@{}}7\\ ($\pm$1.1)\end{tabular}} &
  \multicolumn{1}{c|}{\cellcolor[HTML]{FFFFFF}\begin{tabular}[c]{@{}c@{}}4.3\\ ($\pm$0.6)\end{tabular}} &
  \multicolumn{1}{c|}{\cellcolor[HTML]{FFFFFF}\begin{tabular}[c]{@{}c@{}}10.3\\ ($\pm$0.9)\end{tabular}} &
  \cellcolor[HTML]{FFFFFF}\begin{tabular}[c]{@{}c@{}}7.3\\ ($\pm$2.2)\end{tabular} &
  \multicolumn{1}{c|}{\cellcolor[HTML]{EFEFEF}\begin{tabular}[c]{@{}c@{}}10.5\\ ($\pm$1.6)\end{tabular}} &
  \multicolumn{1}{c|}{\cellcolor[HTML]{FFFFFF}\begin{tabular}[c]{@{}c@{}}8.1\\ ($\pm$1)\end{tabular}} &
  \multicolumn{1}{c|}{\cellcolor[HTML]{FFFFFF}\begin{tabular}[c]{@{}c@{}}4.7\\ ($\pm$0.5)\end{tabular}} &
  \multicolumn{1}{c|}{\cellcolor[HTML]{FFFFFF}\begin{tabular}[c]{@{}c@{}}12.1\\ ($\pm$0.5)\end{tabular}} &
  \cellcolor[HTML]{FFFFFF}\begin{tabular}[c]{@{}c@{}}10.5\\ ($\pm$1.7)\end{tabular} \\ \hline
\textbf{\begin{tabular}[c]{@{}l@{}}Global\\ Iterations\end{tabular}} &
  \multicolumn{1}{c|}{\cellcolor[HTML]{EFEFEF}\begin{tabular}[c]{@{}c@{}}15.8\\ ($\pm$3.7)\end{tabular}} &
  \multicolumn{1}{c|}{\cellcolor[HTML]{FFFFFF}\begin{tabular}[c]{@{}c@{}}15.4\\ ($\pm$3.8)\end{tabular}} &
  \multicolumn{1}{c|}{\cellcolor[HTML]{FFFFFF}\begin{tabular}[c]{@{}c@{}}15.5\\ ($\pm$4)\end{tabular}} &
  \multicolumn{1}{c|}{\cellcolor[HTML]{FFFFFF}\begin{tabular}[c]{@{}c@{}}15.3\\ ($\pm$3.9)\end{tabular}} &
  \cellcolor[HTML]{FFFFFF}\begin{tabular}[c]{@{}c@{}}15.6\\ ($\pm$4)\end{tabular} &
  \multicolumn{1}{c|}{\cellcolor[HTML]{EFEFEF}\begin{tabular}[c]{@{}c@{}}17.9\\ ($\pm$2.7)\end{tabular}} &
  \multicolumn{1}{c|}{\cellcolor[HTML]{FFFFFF}\begin{tabular}[c]{@{}c@{}}17.5\\ ($\pm$2.6)\end{tabular}} &
  \multicolumn{1}{c|}{\cellcolor[HTML]{FFFFFF}\begin{tabular}[c]{@{}c@{}}17.3\\ ($\pm$2.6)\end{tabular}} &
  \multicolumn{1}{c|}{\cellcolor[HTML]{FFFFFF}\begin{tabular}[c]{@{}c@{}}17.6\\ ($\pm$2.4)\end{tabular}} &
  \cellcolor[HTML]{FFFFFF}\begin{tabular}[c]{@{}c@{}}18\\ ($\pm$3.1)\end{tabular} &
  \multicolumn{1}{c|}{\cellcolor[HTML]{EFEFEF}\begin{tabular}[c]{@{}c@{}}18.2\\ ($\pm$2.2)\end{tabular}} &
  \multicolumn{1}{c|}{\cellcolor[HTML]{FFFFFF}\begin{tabular}[c]{@{}c@{}}18\\ ($\pm$2)\end{tabular}} &
  \multicolumn{1}{c|}{\cellcolor[HTML]{FFFFFF}\begin{tabular}[c]{@{}c@{}}17.8\\ ($\pm$2.1)\end{tabular}} &
  \multicolumn{1}{c|}{\cellcolor[HTML]{FFFFFF}\begin{tabular}[c]{@{}c@{}}18.2\\ ($\pm$2.2)\end{tabular}} &
  \cellcolor[HTML]{FFFFFF}\begin{tabular}[c]{@{}c@{}}18\\ ($\pm$2.1)\end{tabular} \\ \hline
\end{tabular}%
}
\vspace{-2pt}
\caption{\textbf{Dynamic Environment - MNIST}: Comparison of our solution (SAC) against four baseline schedulers}
\label{tab:baseline_comparison_dynamic}
\vspace{-12pt}
\end{table*}
The second set of results includes the evaluation of the performance and generalization capabilities of the trained SAC model at the online inference phase for both network environments (static and dynamic) and FL-CNN architectures (MNIST and CIFAR-10). It should be noted that all statistical results are averaged, including also their standard deviation ($\pm$STD), over 100 independent real FL executions (no simulated FL environment is used), in order to draw accurate conclusions. Firstly, a comparison between our proposed solution and four baseline schemes is performed, for the 5, 10 and 20 workers. We compare our SAC DRL solution against a Best Effort, a Random Selection, a Greedy Selection \cite{9384231Zhan}, as well as a Genetic Algorithm Scheduler. 
\begin{itemize}
\item \textbf{Genetic Algorithm Scheduler (GAS): } The GAS is an advanced hybrid genetic meta-heuristic algorithm, and is based on our previous work introduced in \cite{magoula2023safe}.
\item \textbf{Best Effort Scheduler (BES): } The BES selects in each \textit{global} iteration of an FL process, the maximum available capacities of each worker, without considering the energy aspect of the system. The objective of such scheduler is the acceleration of the FL process.
\item \textbf{Random Selection Scheduler (RSS): } In each \textit{global} iteration, the RSS orchestrates randomly the resources of each worker, based on their available capacities. 
\item \textbf{Greedy Selection Scheduler (GSS): } In each \textit{global} iteration, the GSS chooses the resource capacities of all workers that led to the best outcome so far, in terms of the total energy consumption. 
\end{itemize}

Tables \ref{tab:baseline_comparison_static} and \ref{tab:baseline_comparison_dynamic} include the performance comparison of our solution, marked as SAC, against the baseline schedulers under different network environments using MNIST FL-CNN, with regard to: \textbf{1)} the total energy consumption in Joules  (sum of the total computation and transmission energy), 
 \textbf{2)} the total training duration of each \textit{global} iteration in seconds, and \textbf{3)} the total number of \textit{global} iterations. As it can be deduced from the tables, our proposed solution outperforms all four baseline schedulers, in terms of the energy aspect, regardless of the number of total workers and the type of network environment. 

\textbf{Static Environment:}
Table \ref{tab:baseline_comparison_static} shows that our solution for the case of 5 workers, achieves a significant reduction in the total energy consumption of [84, 86, 69] \%, compared to the BES, RSS and GSS, respectively. Similarly, for the case of 10 and 20 workers, our solution results in a similar percentage decrease of [94, 86, 78] \% and [93, 83, 75] \%, respectively. Our approach is capable of improving upon the already energy efficient performance of our GAS approach, achieving a notable decrease in the overall energy consumption of [18, 32, 28] \%. Finally, it should be noted that compared to the BES, which is the fastest option in terms of FL execution, our solution achieves on average 17.2 times lower energy consumption for the different number of workers. However, the trade-off for achieving such considerable reduction, is a slower convergence speed. As it appears in the total training time for all \textit{global} iterations of the FL process, our solution is on average 3.42 times slower compared to the BES for all different workers. 

\textbf{Dynamic Environment:}
Table \ref{tab:baseline_comparison_dynamic} shows an overall increase in the energy consumption for both our solution and the four baseline schedulers compared to the static environment. Intuitively, as the dynamicity of the network increases, so does the complexity towards an optimal strategy. However, our solution once again showcases a robust performance by achieving a significant reduction in the total energy consumption up to 94\% compared to the four baselines, as in the case of static environment. Once again, our SAC-based solution builds upon the already energy efficient GAS approach, proving to be 34\% more energy efficient. Finally, the trade-off discussed in the static environment between FL convergence speed and energy consumption remains, where our solution is on average 2.16 times slower, while consuming 15.72 times lower energy, for the different number of workers.

\textbf{FL Architectures:}
As already stated in Section \ref{simulated_FL}, our SAC DRL model is trained using a simulated FL environment based on the MNIST FL-CNN (depicted as SAC (MNIST FL-CNN) for this section only). However, to evaluate the robustness of our solution not only on different network conditions but also on different FL architectures, we use the more complex and unknown (to SAC) CIFAR-10 FL-CNN on the most challenging scenario, i.e. the dynamic network environment with 20 workers. To examine the sensitivity of our policy in this unknown scenario, we performed an additional comparison analysis against a dedicated SAC DRL model (depicted as SAC (CIFAR-10 FL-CNN)), trained explicitly on the CIFAR-10 FL-CNN. It should be noted that since the CIFAR-10 FL-CNN is a much more complex architecture, the amount of consumed energy is expressed in kJ. The quantitative results of Table \ref{tab:baseline_comparison_dynamic_cifar} demonstrate that the SAC policy exhibits a high degree of tolerance to task changes, leading to a comparable amount of consumed energy (+14.8\%) against the SAC (CIFAR-10 FL-CNN) policy. 
\begin{table}[h]
\centering
\resizebox{\columnwidth}{!}{%
\begin{tabular}{|l|cccccc|}
\hline
\multicolumn{1}{|c|}{\textbf{\begin{tabular}[c]{@{}c@{}}Total Avg.\\ ($\pm$STD)\end{tabular}}} & \multicolumn{6}{c|}{\textbf{20 Workers}} \\ \hline
\textbf{Schedulers} & \multicolumn{1}{c|}{\cellcolor[HTML]{EFEFEF}\textbf{\begin{tabular}[c]{@{}c@{}}SAC\\ (MNIST \\ FL-CNN)\end{tabular}}} & \multicolumn{1}{c|}{\cellcolor[HTML]{EFEFEF}\textbf{\begin{tabular}[c]{@{}c@{}}SAC\\ (CIFAR-10 \\ FL-CNN)\end{tabular}}} & \multicolumn{1}{c|}{\textbf{GAS}} & \multicolumn{1}{c|}{\textbf{BES}} & \multicolumn{1}{c|}{\textbf{RSS}} & \textbf{GSS} \\ \hline
\textbf{\begin{tabular}[c]{@{}l@{}}Total \\ Energy (kJ)\end{tabular}} & \multicolumn{1}{c|}{\cellcolor[HTML]{EFEFEF}\begin{tabular}[c]{@{}c@{}}32.4\\ ($\pm$12)\end{tabular}} & \multicolumn{1}{c|}{\cellcolor[HTML]{EFEFEF}\begin{tabular}[c]{@{}c@{}}27.6\\ ($\pm$9.7)\end{tabular}} & \multicolumn{1}{c|}{\begin{tabular}[c]{@{}c@{}}59\\ ($\pm$39.2)\end{tabular}} & \multicolumn{1}{c|}{\begin{tabular}[c]{@{}c@{}}430.8\\ ($\pm$278.2)\end{tabular}} & \multicolumn{1}{c|}{\begin{tabular}[c]{@{}c@{}}145.5\\ ($\pm$69.5)\end{tabular}} & \begin{tabular}[c]{@{}c@{}}125.2\\ ($\pm$69.9)\end{tabular} \\ \hline
\textbf{\begin{tabular}[c]{@{}l@{}}Computation\\ Energy (kJ)\end{tabular}} & \multicolumn{1}{c|}{\cellcolor[HTML]{EFEFEF}\begin{tabular}[c]{@{}c@{}}32.3\\ ($\pm$11.9)\end{tabular}} & \multicolumn{1}{c|}{\cellcolor[HTML]{EFEFEF}\begin{tabular}[c]{@{}c@{}}27.5\\ ($\pm$9.7)\end{tabular}} & \multicolumn{1}{c|}{\begin{tabular}[c]{@{}c@{}}58.2\\ ($\pm$39)\end{tabular}} & \multicolumn{1}{c|}{\begin{tabular}[c]{@{}c@{}}425.7\\ ($\pm$277.6)\end{tabular}} & \multicolumn{1}{c|}{\begin{tabular}[c]{@{}c@{}}142.9\\ ($\pm$69.1)\end{tabular}} & \begin{tabular}[c]{@{}c@{}}122.2\\ ($\pm$69.7)\end{tabular} \\ \hline
\textbf{\begin{tabular}[c]{@{}l@{}}Transmission\\ Energy (kJ)\end{tabular}} & \multicolumn{1}{c|}{\cellcolor[HTML]{EFEFEF}\begin{tabular}[c]{@{}c@{}}0.1\\ ($\pm$0)\end{tabular}} & \multicolumn{1}{c|}{\cellcolor[HTML]{EFEFEF}\begin{tabular}[c]{@{}c@{}}0.1\\ ($\pm$0)\end{tabular}} & \multicolumn{1}{c|}{\begin{tabular}[c]{@{}c@{}}0.8\\ ($\pm$0.3)\end{tabular}} & \multicolumn{1}{c|}{\begin{tabular}[c]{@{}c@{}}5.1\\ ($\pm$1.5)\end{tabular}} & \multicolumn{1}{c|}{\begin{tabular}[c]{@{}c@{}}2.6\\ ($\pm$0.8)\end{tabular}} & \begin{tabular}[c]{@{}c@{}}3\\ ($\pm$0.7)\end{tabular} \\ \hline
\textbf{\begin{tabular}[c]{@{}l@{}}Training Time \\ per Global \\ Iteration (s)\end{tabular}} & \multicolumn{1}{c|}{\cellcolor[HTML]{EFEFEF}\begin{tabular}[c]{@{}c@{}}765.7\\ ($\pm$111.6)\end{tabular}} & \multicolumn{1}{c|}{\cellcolor[HTML]{EFEFEF}\begin{tabular}[c]{@{}c@{}}814.4\\ ($\pm$91.3)\end{tabular}} & \multicolumn{1}{c|}{\begin{tabular}[c]{@{}c@{}}439.8\\ ($\pm$217.2)\end{tabular}} & \multicolumn{1}{c|}{\begin{tabular}[c]{@{}c@{}}200.3\\ ($\pm$143)\end{tabular}} & \multicolumn{1}{c|}{\begin{tabular}[c]{@{}c@{}}745.7\\ ($\pm$68.2)\end{tabular}} & \begin{tabular}[c]{@{}c@{}}792.7\\ ($\pm$136.4)\end{tabular} \\ \hline
\textbf{\begin{tabular}[c]{@{}l@{}}Global\\ Iterations\end{tabular}} & \multicolumn{1}{c|}{\cellcolor[HTML]{EFEFEF}\begin{tabular}[c]{@{}c@{}}85.5\\ ($\pm$17.1)\end{tabular}} & \multicolumn{1}{c|}{\cellcolor[HTML]{EFEFEF}\begin{tabular}[c]{@{}c@{}}62.9\\ ($\pm$7.8)\end{tabular}} & \multicolumn{1}{c|}{\begin{tabular}[c]{@{}c@{}}78.8\\ ($\pm$17.9)\end{tabular}} & \multicolumn{1}{c|}{\begin{tabular}[c]{@{}c@{}}90\\ ($\pm$25.4)\end{tabular}} & \multicolumn{1}{c|}{\begin{tabular}[c]{@{}c@{}}101.5\\ ($\pm$26.9)\end{tabular}} & \begin{tabular}[c]{@{}c@{}}109.6\\ ($\pm$14.8)\end{tabular} \\ \hline
\end{tabular}%
}
\caption{\textbf{Dynamic Environment - CIFAR-10}: Comparison of our solution (SAC (MNIST FL-CNN) \& SAC (CIFAR-10 FL-CNN)) against four baseline schedulers}
\label{tab:baseline_comparison_dynamic_cifar}
\vspace{-5pt}
\end{table}
These results suggest that a single, robust and generalizable DRL model is sufficient to cover different tasks of varying complexity, whereas in the opposite case additional training or re-training of dedicated models would be required, wasting significant resources (e.g. hyperparameter tuning, convergence speed depending on the task). Regarding the four baselines, our solution once again manages to significantly reduce the total amount of consumed energy.  More specifically, our solution achieves a decrease equal to [45, 92, 77, 74] \% compared to the GAS, BES, RSS and GSS, respectively. Remarkably, the performance of our solution on a challenging and unknown evaluation scenario, validates our claims towards a robust and generalized SAC DRL approach.

To summarize, regardless of the complexity of the problem, our solution results in an energy efficient FL, providing reasonable trade-off between energy consumption and training time, vastly outperforming the three state-of-the-art baseline schedulers (BES, RSS, GSS) and further improving upon the already energy efficient hybrid genetic meta-heuristic algorithm (GAS). Our solution demonstrates high tolerance on different FL tasks, showcasing the flexibility, robustness and generalization capabilities of our proposed solution.  

\subsection{Worker Side Synchronization} 
\vspace{-2pt}
The last set of evaluation results refers to our proposed worker side synchronization method compared to the traditional one that takes place at the coordinator side \cite{Jiang9780218}. The primary difference between the two approaches relies on the responsibility for discarding any late model updates. Contrary to our method, in the coordinator side synchronization, the coordinator is in charge of discarding any late model updates. 
It should be noted that all results were extracted using the MNIST FL-CNN in the case of 20 workers in the dynamic network environment. Table \ref{tab:sync_results} summarizes the gains of our proposed method for different time thresholds, in terms of: 
 \textbf{1)} total wasted energy, \textbf{2)} total unnecessary channel accesses (late model update transmissions) and \textbf{3)} the respective channel occupation time for a complete FL process. Intuitively the stricter the time threshold the higher the amount of wasted energy, as a higher number of workers are unable to complete their FL tasks. In our synchronization method both the unnecessary channel accesses and channel occupation time are eliminated. Focusing on the coordinator synchronization, more unnecessary channel accesses lead to higher channel occupancy time. Especially in the case of $\mathsf{H}$ equal to 6 seconds, the involved workers are accessing the channel 53.5 times occupying it for 43.3 seconds on average, wasting communication resources. Once again, this highlights the merits of our synchronization method.
\vspace{-3pt}
\begin{table}[h!]
\centering
\resizebox{\columnwidth}{!}{%
\begin{tabular}{|l|
>{\columncolor[HTML]{EFEFEF}}c 
>{\columncolor[HTML]{EFEFEF}}c 
>{\columncolor[HTML]{EFEFEF}}c |ccc|}
\hline
\multicolumn{1}{|c|}{\textbf{\begin{tabular}[c]{@{}c@{}}Avg.Value\\ ($\pm$STD)\end{tabular}}} &
  \multicolumn{3}{c|}{\cellcolor[HTML]{EFEFEF}\textbf{Worker Side Sync}} &
  \multicolumn{3}{c|}{\textbf{Coordinator Side Sync}} \\ \hline
\textbf{\begin{tabular}[c]{@{}l@{}}Time \\ Threshold\end{tabular}} &
  \multicolumn{1}{c|}{\cellcolor[HTML]{EFEFEF}\textbf{13 s}} &
  \multicolumn{1}{c|}{\cellcolor[HTML]{EFEFEF}\textbf{8 s}} &
  \textbf{6 s} &
  \multicolumn{1}{c|}{\textbf{13 s}} &
  \multicolumn{1}{c|}{\textbf{8 s}} &
  \textbf{6 s} \\ \hline
\textbf{\begin{tabular}[c]{@{}l@{}}Wasted\\ Energy (J)\end{tabular}} &
  \multicolumn{1}{c|}{\cellcolor[HTML]{EFEFEF}\begin{tabular}[c]{@{}c@{}}0.6\\ ($\pm$0.9)\end{tabular}} &
  \multicolumn{1}{c|}{\cellcolor[HTML]{EFEFEF}\begin{tabular}[c]{@{}c@{}}8.3\\ ($\pm$6.1)\end{tabular}} &
  \begin{tabular}[c]{@{}c@{}}25.4\\ ($\pm$11.9)\end{tabular} &
  \multicolumn{1}{c|}{\begin{tabular}[c]{@{}c@{}}1\\ ($\pm$1.9)\end{tabular}} &
  \multicolumn{1}{c|}{\begin{tabular}[c]{@{}c@{}}9.8\\ ($\pm$7.6)\end{tabular}} &
  \begin{tabular}[c]{@{}c@{}}29.5\\ ($\pm$17.3)\end{tabular} \\ \hline
\textbf{\begin{tabular}[c]{@{}l@{}}Unnecessary\\ Channel \\ Accesses\end{tabular}} &
  \multicolumn{1}{c|}{\cellcolor[HTML]{EFEFEF}-} &
  \multicolumn{1}{c|}{\cellcolor[HTML]{EFEFEF}-} &
  - &
  \multicolumn{1}{c|}{\begin{tabular}[c]{@{}c@{}}2.9\\ ($\pm$4.7)\end{tabular}} &
  \multicolumn{1}{c|}{\begin{tabular}[c]{@{}c@{}}20.9\\ ($\pm$15.5)\end{tabular}} &
  \begin{tabular}[c]{@{}c@{}}53.5\\ ($\pm$25.3)\end{tabular} \\ \hline
\textbf{\begin{tabular}[c]{@{}l@{}}Unnecessary\\ Channel\\ Occupation \\ Time (s)\end{tabular}} &
  \multicolumn{1}{c|}{\cellcolor[HTML]{EFEFEF}-} &
  \multicolumn{1}{c|}{\cellcolor[HTML]{EFEFEF}-} &
  - &
  \multicolumn{1}{c|}{\begin{tabular}[c]{@{}c@{}}4.4\\ ($\pm$9.3)\end{tabular}} &
  \multicolumn{1}{c|}{\begin{tabular}[c]{@{}c@{}}23.5\\ ($\pm$17.1)\end{tabular}} &
  \begin{tabular}[c]{@{}c@{}}43.3\\ ($\pm$16.5)\end{tabular} \\ \hline
\end{tabular}%
}
\caption{Worker vs Coordinator Side Synchronization}
\label{tab:sync_results}
\vspace{-10pt}
\end{table}
\vspace{-4pt}
\section{Conclusions}\label{conclusions}
\vspace{-2pt}
This paper proves the effectiveness and feasibility of the proposed safe DRL solution, targeting the minimization of
the overall energy consumption of an FL process in a wireless
communication network. A simulated
and significantly simpler FL environment was designed for
the SAC training phase, emulating a real FL process, targeting a lower complexity and faster convergence. 
The decision making of the SAC agent advances, through the introduction of a penalty function that almost minimizes the number of constraint violations, as it can be deduced from the evaluation section. Evaluation results include 100 real FL independent runs, under different network environments (static and dynamic), and different FL architectures (MNIST and CIFAR-10 FL-CNNs). For the offline phase, our solution achieves an energy consumption decrease of up to 82.7\%. For the online phase, comparisons are performed against four state-of-the-art baselines, where our solution achieves a decrease of up to 94\% in the total energy consumption. Evaluation results on a challenging and unknown scenario are also included, validating our claims towards a robust and generalized SAC DRL approach. Finally, we propose a worker side synchronization that compared to the traditional one, eliminates the unnecessary channel accesses and total channel occupancy time. 
The future work includes an investigation for an energy efficient FL approach, considering also renewable energy sources, towards green communications. Additional types of resources related to the overall energy consumption will be also examined, in order to further improve the carbon footprint of FL in wireless communication networks.

\bibliographystyle{IEEEtran}
\vspace{-4pt}
\bibliography{bibliography/bib}

\end{document}